\documentclass[runningheads]{llncs}

\usepackage{eccv}

\usepackage{eccvabbrv}

\usepackage{graphicx}
\usepackage{booktabs}
\usepackage{wrapfig}
\usepackage[ruled,vlined]{algorithm2e}
\usepackage{tcolorbox}
\definecolor{verylightgray}{gray}{0.9}
\usepackage{placeins}

\usepackage[accsupp]{axessibility}  

\usepackage{hyperref}

\usepackage{orcidlink}

\begin{document}

\title{``World Knowledge'' in the Weights: Reading Concept Circuits of Vision Transformers} 


\author{Yanlin Chen\inst{1}\orcidlink{0009-0004-9382-5788} \and
Tang Li\inst{1}\orcidlink{0000-0002-3134-4151} \and
Xi Peng\inst{2,\dagger}\orcidlink{0000-0002-7772-001X}}

\authorrunning{Y.~Chen et al.}

\institute{Department of Computer \& Information Sciences, University of Delaware, USA
\email{\{yanlin,tangli\}@udel.edu} \and
Department of Computer Science, University of Virginia, USA\\
\email{naq5rd@virginia.edu}
}

\renewcommand{\thefootnote}{}
\footnotetext[1]{$^\dagger$ Corresponding author.}

\maketitle

\begin{abstract}
Vision transformers (ViTs) have achieved remarkable generalization across visual domains, yet little is known about how they internally represent the structure of the world. To address this gap, we use Cross-Layer Transcoders (CLTs) to read \emph{concept circuits} from ViTs: directed graphs whose nodes correspond to sparse, interpretable concepts and edges capture concept interactions across layers. Our method yields two complementary views of model behavior. The \emph{global} concept circuit is input-invariant and can be recovered directly from learned cross-layer weights, exposing the reusable “world knowledge” encoded in the model. The \emph{instance} concept circuit is input-dependent and identifies the concepts and pathways actually used for a specific prediction, enabling faithful example-level explanations. We demonstrate the utility of concept circuits in three ways: (1) Automatic spurious correlation discovery: leveraging the statistics of our global concept circuits to identify shortcut dependencies within the model. (2) Spurious correlation removal: intervening on the instance concept circuit to steer the model towards correct predictions. Empirical results show that our method outperforms existing counterparts by 11.0\% on the Waterbird dataset. (3) Model comparison: contrasting the global concept circuits of different foundation models (e.g., CLIP vs. DINO) to reveal how supervision paradigms shape representational structure. Our code is available at \url{https://github.com/deep-real/VisionCLT}

\end{abstract}    
\section{Introduction}
\label{sec:intro}
Vision transformers (ViTs)~\cite{dosovitskiy2021an} have achieved remarkable generalization across a wide range of visual domains~\cite{wang2021not,chen2023vision,wu2024clipself,radford2021learning}.
Despite this empirical success, we still lack an understanding of the inner workings of ViTs that give rise to robust behavior or well-known failures such as reliance on spurious correlations~\cite{ye2024spurious,xiao2021noise,moayeri2022comprehensive,geirhos2018imagenet,peng2026inside}, \ie their \textbf{``world knowledge''}. In particular, it remains unclear what concepts ViTs encode and how these concepts interact across layers.
Recovering this internal structure is crucial because it enables principled auditing, diagnosis of shortcuts, and targeted interventions at the level of internal mechanisms.

Yet, reading the ``world knowledge'' of ViTs remains challenging. Most existing interpretable machine learning (IML) methods focus on saliency maps~\cite{simonyan2013deep,zhou2016learning,selvaraju2017grad} or feature-importance scores~\cite{ribeiro2016should,lundberg2017unified}, highlighting which input regions or features matter for a prediction but treating the network’s internals as a black box.
\begin{wrapfigure}{r}{0.45\textwidth}
  \centering
  \fboxsep=0pt 
  \includegraphics[width=\linewidth]{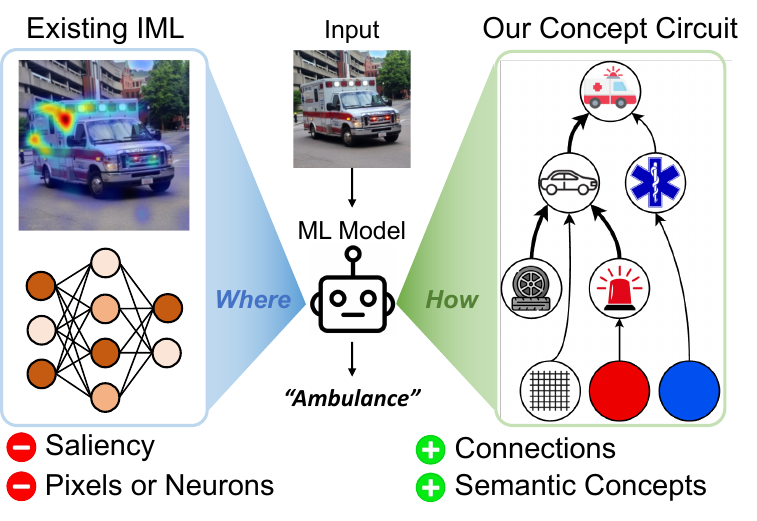}
  \caption{A comparison between existing {\it Interpretable Machine Learning (IML)} and our \textit{concept circuit}. Existing IML methods are typically limited to {\it where} the model focuses on, while our concept circuit answers the question of {\it how} the model understands the visual world.}
  \label{fig:title_figure}
  \vspace{-18pt}
\end{wrapfigure}While recent Sparse Autoencoder (SAE) methods~\cite{huben2023sparse,bricken2023monosemanticity,rajamanoharan2024jumping,gao2025scaling} decompose polysemantic neurons into sparse, more interpretable features and support circuit discovery~\cite{neel2022attribution,conmy2023towards,kramar2024atp} to identify ``sparse feature circuits''~\cite{marks2025sparse}, SAE-based pipelines primarily capture \textit{input-dependent} behavior~\cite{dunefsky2024transcoders}, which does not reflect the model's general behavior. Although one can obtain a global view of feature interactions by averaging over large datasets, it requires expensive attribution and still risks being not fully input-invariant.

To address these gaps, we propose to use \emph{Cross-Layer Transcoders} (CLTs)~\cite{ameisen2025circuit} to read the ``world knowledge'' of ViTs (Fig.~\ref{fig:title_figure}). CLTs are sparse cross-layer dictionaries trained to approximate the computation performed by language transformers. Specifically, they encode the residual stream into sparse feature activations and use a learned decoder to reconstruct the downstream MLP outputs across layers.
Compared to SAEs, a key advantage of CLTs is that they learn \textit{input-invariant} encoder/decoder weights which define how features interact across layers in an input-independent manner. 
By adapting CLTs to vision data, we recover \textbf{global concept circuits} directly from learned parameters without expensive dataset-scale attribution. This enables systematic analyses of model's general behavior and principled auditing of model bias. 
Additionally, CLTs also decompose model representation into input-dependent sparse codes, which we leverage to extract \textbf{instance concept circuits}. This offers a view complementary to global concept circuits, enabling fine-grained auditing and control at instance level. Although SAEs can also identify instance concept circuits, we show that CLT-based circuits are more faithful than SAE-based circuits.

We showcase the utility of concept circuits through three key applications:
(1) \textbf{Spurious correlation discovery}: The global concept circuits allow us to \textit{automatically} identify shortcut dependencies; to our knowledge, this is the first work that aims to uncover spurious correlations without domain knowledge.   
(2) \textbf{Spurious correlation removal}: Leveraging the instance-specific concept circuits, we can localize the specific concepts/interactions responsible for biased behavior, and intervene to reduce the model’s reliance on spurious cues. Empirical results show that our method outperforms existing counterparts by 11.0\% on the Waterbird dataset.
(3) \textbf{Model comparison}: By analyzing the structure of global concept circuits across different pretrained foundation models ({\it e.g.}, CLIP~\cite{radford2021learning} v.s. DINO~\cite{caron2021emerging} v.s. ViT~\cite{dosovitskiy2021an}), we can evaluate and compare how various supervision paradigms and datasets influence model behaviors. This provides a principled mechanistic basis for model optimization and selection.

In summary, our contributions are:
\begin{itemize}
    \item We adapt CLTs to ViTs, validating their effectiveness on vision data.
    \item We use CLTs to read instance-specific and, first of its kind, global concept circuits from ViTs, revealing their inner ``world knowledge''.
    \item We showcase the utility of concept circuits in three applications: automatic spurious correlation discovery, spurious correlation removal, where our method outperforms existing counterparts by 11.0\%, and comparing vision foundation models through their internal representation structure.
\end{itemize}

\begin{figure*}[t]
  \centering
  \fboxsep=0pt 
  \includegraphics[width=1.0\linewidth]{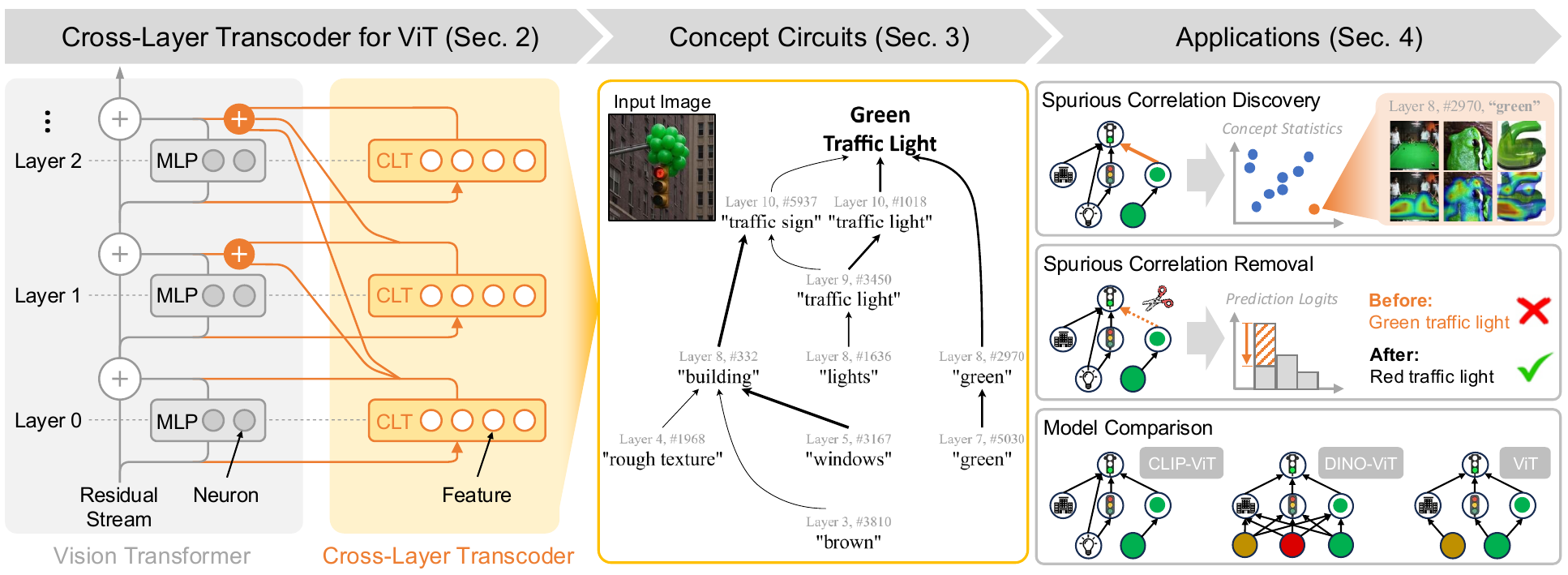}
  \caption{Overview of our pipeline for extracting concept circuits of ViTs. \textbf{Left:} the CLT architecture, which is trained to approximate the MLP computation of a transformer. \textbf{Middle:} The concept circuit extracted by CLT, {\it i.e.,} a graph of concepts and their interactions. \textbf{Right:} The concept circuits can be leveraged to (1) automatically discover spurious correlations encoded in the ViT, (2) remove the spurious correlations to steer the model toward correct predictions, and (3) offer insights on how variant supervision paradigms shape model's internal structure and behavior.}
  \label{fig:overview}
  \vspace{-10pt}
\end{figure*}

\section{Preliminary}
\label{subsec:clt}

\subsection{Sparse Autoencoders (SAEs)}
SAEs~\cite{huben2023sparse,bricken2023monosemanticity,rajamanoharan2024jumping,gao2025scaling} are proposed to decompose the model representations of a \textit{single layer} into interpretable concepts. Applying SAEs to read concept circuits would require training multiple SAEs and integrating with circuit discovery methods~\cite{neel2022attribution,conmy2023towards,kramar2024atp} to quantify the interactions between SAE features~\cite{marks2025sparse}. As discussed in Sec.~\ref{sec:intro}, this approach has two limitations: (1) it primarily captures \emph{input-dependent} interactions between features, and (2) it requires large scale attribution over datasets to get a global view of feature interactions, which is computationally expensive. These limitations hinder SAE's potential for reading \emph{global} concept circuits.

\subsection{Cross-Layer Transcoders (CLTs)}
CLTs~\cite{ameisen2025circuit} are sparse \emph{cross-layer} dictionaries trained to approximate Multi-Layer Perceptron (MLP) computation of a transformer.
Concretely, CLT encodes the MLP input $\mathbf{x}$ at one layer to sparse hidden activations $\mathbf{z}$ and reconstructs the MLP output $\mathbf{y}$ using all hidden activations at previous layers. Let $\mathbf{W}_{\mathrm{enc}}^l\in\mathbb{R}^{f\times d}$ denote the encoder for layer $l$, $\mathbf{W}_{\mathrm{dec}}^{l'\to l}\in\mathbb{R}^{d\times f}$ denote the decoder for layer $l'$ writing to layer $l$, and $\phi$ denote the activation function to impose sparsity, a CLT can be expressed as:
\begin{equation}
\mathbf{z}^l=\mathrm{\phi}(\mathbf{W}_{\mathrm{enc}}^l\mathbf{x}^l),
\quad
\hat{\mathbf{y}}^l=\sum_{l'=1}^l\mathbf{W}_{\mathrm{dec}}^{l'\to l}\mathbf{z}^{l'}.
\end{equation}
The weights for all layers are jointly optimized by minimizing the reconstruction error of all layers with a sparsity penalty: $\mathcal{L}_{\mathrm{CLT}}=\sum_{l=1}^L\|\mathbf{y}^l-\hat{\mathbf{y}}^l\|_2^2+\lambda\mathcal{L}_{\mathrm{spa}}.$
By constraining sparsity of the hidden activations, CLTs are enforced to encode disentangled, human interpretable concepts.

\vspace{1em}
\noindent\textbf{Advantages of CLTs.}
As discussed before, combining SAEs and circuit discovery methods has limitations in reading \emph{global} concept circuits. In contrast, CLT factorizes the original model's computations into two parts: (1) \emph{input-dependent} sparse codes $\mathbf{z}$, and (2) \emph{input-invariant} encoder/decoder weights, which are fixed once the CLT is trained and do not change over inputs. This factorization enables two key modes of analysis. First, the sparse codes provide faithful feature activation on specific inputs, supporting fine-grained attribution for reading instance concept circuits. Second, the learned weights encode a global view of concept interactions in a input-independent manner, allowing for direct readout of global concept circuits.

\section{Method}
While ViTs have achieved great success across a wide range a visual domains, their inner workings remain a black box. This undermines their reliability in safety-critical domains, such as medical image analysis and autonomous driving. 
In this paper, we propose to open the black box, reading ViT's ``world knowledge'': structured graphs in which nodes are human-interpretable concepts and directed edges capture how concepts interact across layers.
To recover this internal structure, we adapt CLTs to ViTs (\cref{subsec:adapting_clts}) and use them to read concept circuits from ViTs. Specifically, we aim to read two types of concept circuits: global concept circuits (\cref{subsec:global_circuit}) and instance concept circuits (\cref{subsec:instance_circuit}). \cref{fig:overview} shows the overview of our pipeline.

\begin{figure}[t]
  \centering
  \fboxsep=0pt 
  \includegraphics[width=0.8\linewidth]{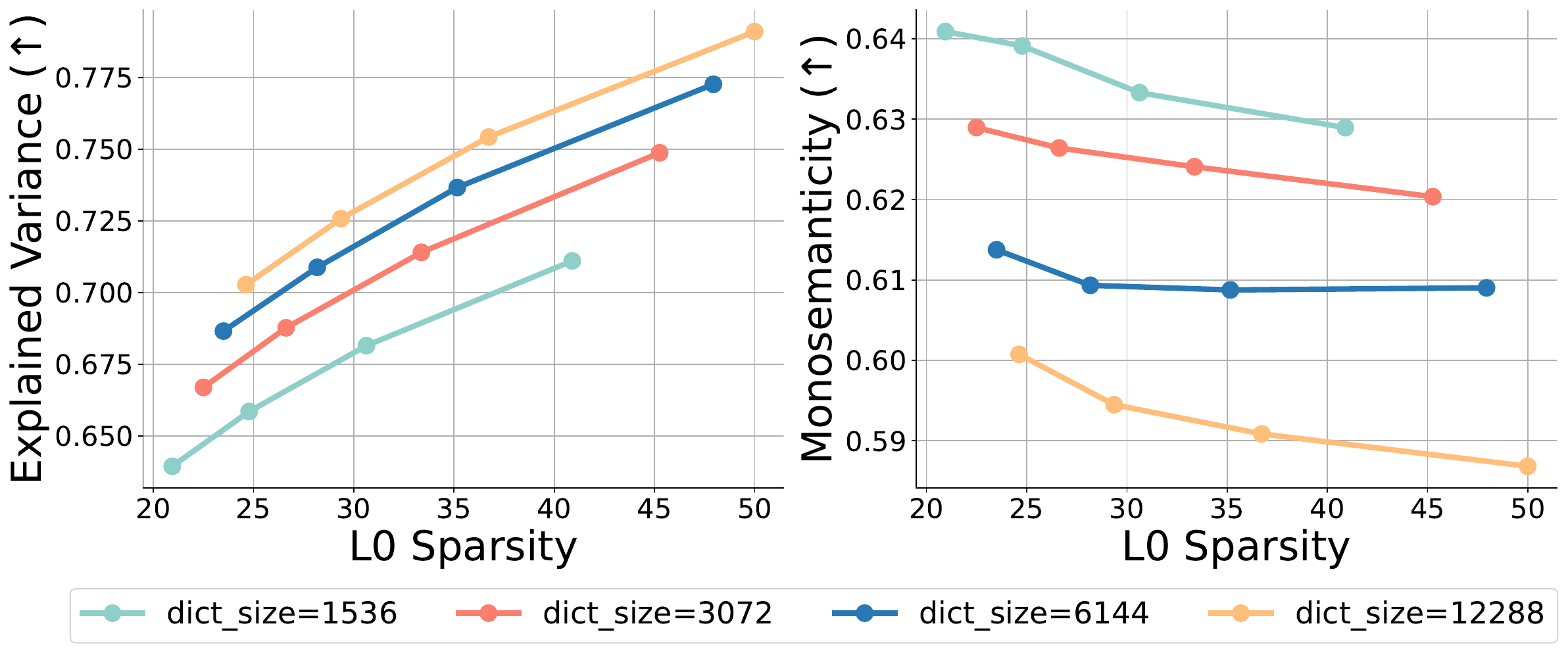}
  \caption{Sweep results of CLT training hyper-parameters. We evaluate on two metrics: \emph{explained variance}, which measures the reconstruction quality, and \emph{monosemanticity}, which measures the interpretability of learned features. Higher $L_0$ sparsity and dictionary size lead to higher explained variance but lower monosemanticity. We select a sparsity of 35 and a dictionary size of 6144 to strike a tradeoff between both metrics.}
  \label{fig:ablation}
  \vspace{-10pt}
\end{figure}
\subsection{Adapting CLTs to ViTs}
\label{subsec:adapting_clts}
CLTs are originally proposed for language models, it is non-trivial to adapt CLTs to ViTs.
There are three major challenges. (1) \emph{Hyper-parameter configuration}. Training CLTs on vision data requires different hyper-parameter choices. To this end, we sweep over a range of hyper-parameter configurations for CLIP CLT on the ImageNet~\cite{deng2009imagenet} training set and evaluate on two metrics: explained variance and monosemanticity~\cite{pach2025sparse}. Based on the evaluation results (Fig.~\ref{fig:ablation}), we choose sparsity of 35 and dictionary size of 6144 for the remaining experiments. Please refer to §\ref{supp:impl_detail} for more implementation details. (2) \emph{Incorporation of \texttt{CLS} tokens}. Different from language models, ViTs often add a \texttt{CLS} token for downstream tasks like classification. Since the \texttt{CLS} tokens contain global information of the image, we choose to train CLTs on image tokens along with \texttt{CLS} tokens. (3) \emph{Simplification of concept circuit}. Naively applying CLTs to ViTs leads to complex concept circuits since ViTs operate on a large number of image tokens ({\it e.g.,} 196 for ViT-B/16 {\it v.s.} 10-20 for language models). Therefore, we choose to aggregate features over token positions to simplify the concept circuit. See §\ref{subsec:instance_circuit} for more details.

To interpret the meaning of a CLT feature, we visualize the top activated images of the feature. To localize the feature in the image, we concatenate the feature activation value of each patch to get an activation map and overlay on the original image. See Fig.~\ref{fig:world_model} for example feature visualizations.

\subsection{Recovering ``world knowledge'' from global concept circuits}
\label{subsec:global_circuit}
Global concept circuits are \textit{input-invariant} circuits induced by the model’s learned parameters, summarizing the concept-to-concept relations the ViT tends to use in general, \eg, a stable pathway where “fur texture” → “four-legged animal” → “dog” frequently appears regardless of the specific image. Reading global concept circuits provides an input-invariant view of the concept-to-concept structure the model tends to reuse, which makes the ViT’s general “world knowledge” explicit and easier to audit. In particular, these circuits support system-level analyses, \eg, automatically detecting shortcut dependencies by comparing global edge strength with co-activation frequency without requiring human annotation.

\vspace{1em}
\noindent\textbf{Method.}
Given a trained CLT, we can read out a global concept circuit directly from its encoder and decoder weights.
Concretely, for two features $s$ and $t$, the global weight between them is the sum of all linear paths inside the CLT, which can be expressed as:

\begin{equation}
w_{s\to t}
=
\Big\langle
\sum_{l \in L_{st}} W^{s,l}_{\mathrm{dec}},
\; W^t_{\mathrm{enc}}
\Big\rangle,    
\label{eq2}
\end{equation}
where $W^{s,l}_{\mathrm{dec}}$ denotes the decoder vector of feature $s$ writing to layer $l$, $W^t_{\mathrm{enc}}$ denotes the encoder vector of feature $t$, and $L_{st}$ is the set of all intermediate layers between $s$ and $t$.

However, as many features are connected through the residual stream, large global weights can arise between features that rarely co-activate, making $w_{s\to t}$ alone a poor descriptor of functional interactions.
To mitigate this feature interference issue and obtain meaningful global edges, we reweight global weights using feature co-activation statistics over the data distribution.
Let $a_i$ be the activation of feature $i$, the reweighted edge weight is
\begin{equation}
\hat{w}_{s\to t}
=
\mathbb{E}\!\left[\mathbf{1}(a_s>0)\,\mathbf{1}(a_t>0)\right] w_{s\to t}.    
\label{eq3}
\end{equation}
These co-activation weighted global weights summarize how strongly feature $s$ tends to influence feature $t$ when they activate at the same time, and we take them as the edge weights of the global concept circuit.

\begin{algorithm}[t]
\caption{Extract Class Subgraphs from Global Concept Circuits}
\label{alg:class_subgraph}
\KwIn{Global circuit $\mathcal{G}=(\mathcal{V},\mathcal{E},\hat{w})$; seed nodes $\mathcal{V}_{\text{seed}}$; branching factor $k$.}
\KwOut{Class subgraph $\mathcal{G}_c=(\mathcal{V}_c,\mathcal{E}_c)$.}

$\mathcal{F} \leftarrow \mathcal{V}_{\text{seed}}$ \tcp*{initialize the frontier}
$\mathcal{V}_c \leftarrow \mathcal{V}_{\text{seed}},\;\; \mathcal{E}_c \leftarrow \emptyset$\;

\While{$\mathcal{F}\neq \emptyset$}{
\tcp{iterative backtracking}
  \ForEach{$u\in \mathcal{V}$}{
    $s(u)\leftarrow \sum_{v\in \mathcal{F}}\hat{w}_{u\rightarrow v}$ \tcp*{total influence to frontier nodes}
  }
  $\mathcal{U} \leftarrow \text{TopK}(\{u\in \mathcal{V} : s(u)>0\}, k)$ \tcp*{filter upstream features}
  \If{$\mathcal{U}=\emptyset$}{\textbf{break}}
  $\mathcal{V}_c \leftarrow \mathcal{V}_c \cup \mathcal{U}$ \tcp*{add nodes to subgraph}
  $\mathcal{E}_c \leftarrow \mathcal{E}_c \cup \{(u\!\to\! v)\in \mathcal{E} : u\in \mathcal{U},\, v\in \mathcal{F}\}$ \tcp*{add edges to subgraph}
  $\mathcal{F} \leftarrow \mathcal{U}$ \tcp*{reinitialize frontier}
}
\Return{$\mathcal{G}_c=(\mathcal{V}_c,\mathcal{E}_c)$}\;
\end{algorithm}

While the full global concept circuit is useful for systematic analysis ({\it e.g.,} automatic spurious correlation discovery), it is too complex for humans to interpret.
Here we introduce how to extract a small class subgraph that is easier to interpret (\cref{alg:class_subgraph}).
Given images of a target class, we run the model with CLTs and select the top-activating last layer features as seed nodes.
Starting from these seeds, we iteratively backtrack the global concept circuit by repeatedly adding a small set of upstream features that have the strongest connections into the current frontier, then treating the newly added features as the next frontier.
We stop when no additional upstream features are selected, and return the induced subgraph as the class subgraph, which contains the features and edges that the ViT uses to represent and reason about a target object.

\vspace{1em}
\begin{table}[tb]
\caption{
Validation of global concept circuit. We mean-ablate top-10 important concepts in the global concept circuits and measure the accuracy drop. We evaluate on 50 randomly selected classes in ImageNet validation set and report mean accuracy.
}
\centering
\resizebox{0.75\linewidth}{!}{%
\begin{tabular}{l@{\hspace{10pt}}c@{\hspace{10pt}}c@{\hspace{10pt}}c@{\hspace{10pt}}c@{\hspace{10pt}}c}
\toprule[0.9pt]
       & Original & Random & Ours & Raw weights & Co-activations \\ \toprule[0.10pt]
CLIP & 0.60 & 0.66 & {\bf 0.22} & 0.48 & 0.50 \\
DINO & 0.76 & 0.73 & {\bf 0.62} & 0.67 & 0.72 \\
ViT & 0.70 & 0.66 & {\bf 0.44} & 0.60 & 0.62 \\
\bottomrule[0.9pt]
\end{tabular}
\label{tab:validation}
}
\end{table}
\noindent\textbf{Validation.}
Although we can directly read out global concept circuits from CLT weights, the validity of the concept structures learned by CLT remains unknown. In this section, we validate the faithfulness of global concept circuits. Specifically, we aim to answer the question: \emph{do the global weights (Eq.~\ref{eq3}) capture genuine concept relations that are encoded in the model and drive the model's decision?}

To answer this question, we conducted a validation experiment on ImageNet validation set: for the global concept circuit of a class, we mean-ablate upstream concepts with top-10 global weights, and measure induced changes in classification accuracy. We test on 50 random classes. We also ablate randomly selected concepts as baseline. To further justify the necessity of reweighting global weights by co-activations, we ablate the effects of only using raw global weights (Eq.~\ref{eq2}) and co-activation statistics to construct the global concept circuit.

Tab.~\ref{tab:validation} shows the validation results on CLIP, DINO and supervised ViT. We find that: (1) \textit{Concepts with high global weight are also functionally important to model's decision.} Mean-ablating top-10 important concepts in the global concept circuits results in significant performance drop in all three models. Notably, the mean accuracy of CLIP drops from $0.60$ to $0.22$. While randomly ablating concepts results in minimal influence on performance. (2) \textit{Co-activation reweighting reduces interference.} The raw global weights (Eq.~\ref{eq2}) may be large between non-related concepts due to feature interference. Experiment results prove this: ablating concepts with top-10 raw weights has less impact than reweighted weights (Eq.~\ref{eq3}). Using co-activation statistics to construct the global concept circuit also leads to suboptimal results, suggesting that the our global weights do not only reflect statistical artifacts, but capture genuine concept relations.

\subsection{Revealing decision making process via instance concept circuits}
\label{subsec:instance_circuit}
Instance concept circuits are \textit{input-dependent} circuits extracted from a single input image, retaining only the concepts and edges that are actually active and influential for a particular prediction, \eg, there may be a pathway of “barking” → “dog” in some dog images, while not in others.
Reading instance concept circuits complements the global view by revealing the actual set of active concepts and interactions responsible for a particular prediction, enabling faithful, instance-level explanations. This is especially useful for fine-grained auditing and control, since it helps localize the specific concepts/interactions driving biased behavior and provides actionable targets for intervention and steering.

\begin{figure*}[t]
  \centering
  \fboxsep=0pt 
  \includegraphics[width=1.0\linewidth]{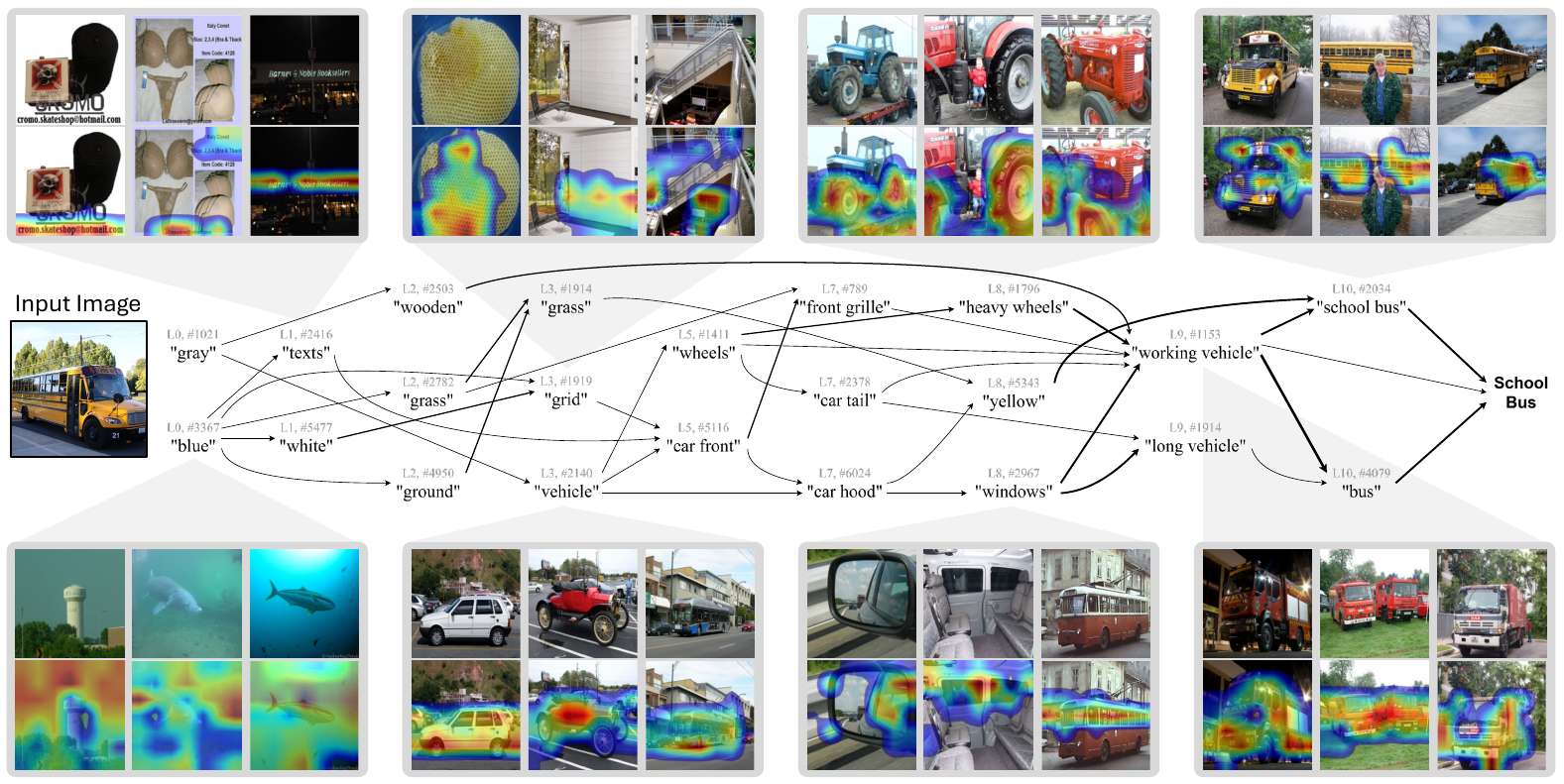}
  \caption{An example instance concept circuit for a ``School Bus'' image and  visualizations of concepts. Thickness of edges represents their importance. The concepts encoded in the model evolve from colors and textures ({\it e.g.,} ``blue'', ``grid'') to object parts ({\it e.g.,} ``wheels'', ``windows''), and finally to objects ({\it e.g.,} ``working vehicle'', ``school bus'').}
  \label{fig:world_model}
\end{figure*}
\vspace{1em}
\noindent\textbf{Method.}
We extract the instance concept circuit from an input image by the following steps: (1) \textit{Calculate attribution weights}. We first run the model with CLTs to get the sparse activations, then use attribution patching~\cite{neel2022attribution} to calculate the attribution weights between each pair of features $A_{s\to t}\;=\;a_s\,\nabla_{a_s}a_t$, where $a_s$ and $a_t$ are the activations of the source and target feature. (2) \textit{Aggregate across token positions}. ViTs have a large number of tokens (\eg, 196 for ViT-B/16), resulting in large circuits that are difficult for human to interpret. Therefore, we aggregate the same feature over different token positions to one node by summing up the associated edge weights. (3) \textit{Prune nodes and edges}. To extract a small, interpretable subgraph, we prune the nodes and edges by their importance and then keep only the most influential ones. Concretely, we first filter nodes with cumulative contribution to the logits above a threshold, then filter edges with high influence to the remaining nodes. This yields a compact subgraph that focuses on the paths most responsible for the model’s prediction.
Fig.~\ref{fig:world_model} demonstrates an example instance concept circuit and feature visualizations of CLIP-ViT-B/32 on a ``School Bus'' image.

\vspace{1em}
\noindent\textbf{Faithfulness evaluation.}
\begin{figure}[t]
  \centering
  \includegraphics[width=0.9\linewidth]{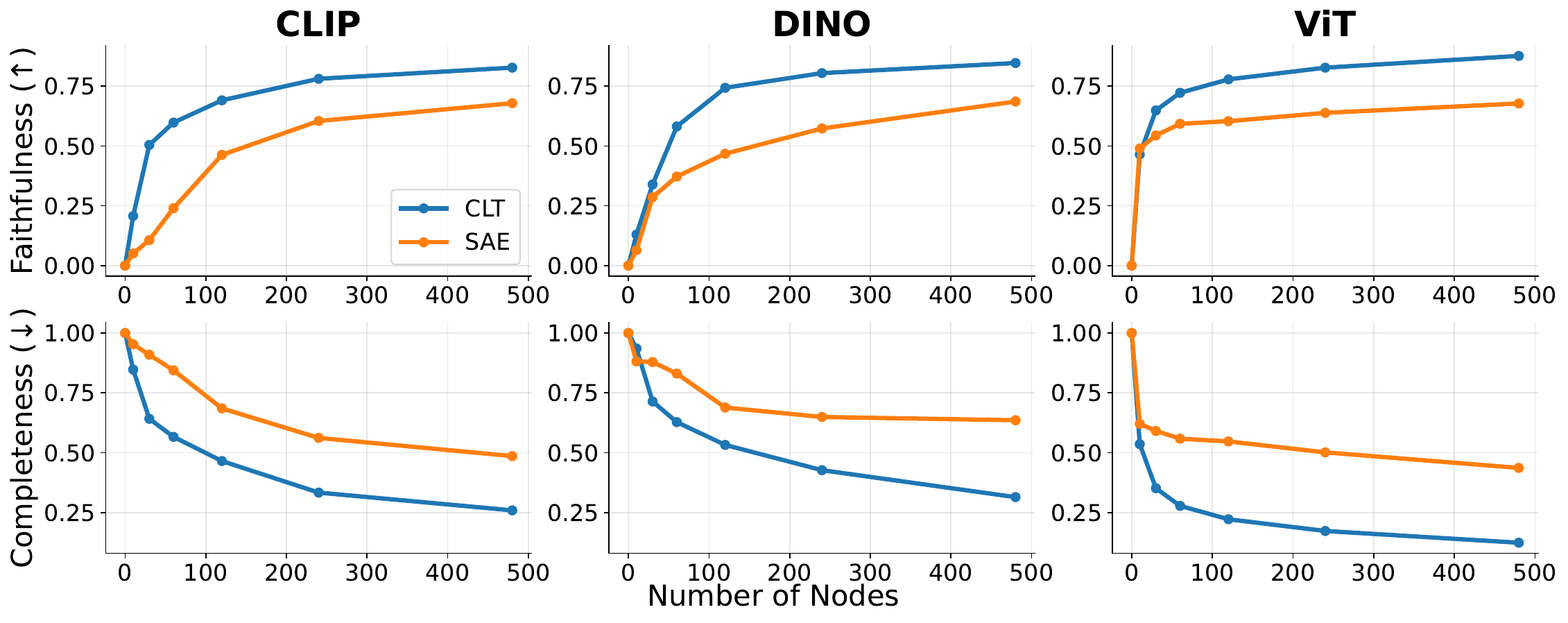}
  \caption{Comparison of faithfulness between SAE-based and CLT-based instance concept circuits on CLIP, DINO and supervised ViT. We evaluate both methods on 50 randomly sampled classes from the ImageNet validation set. CLT consistently outperforms SAE in terms of \emph{faithfulness} and \emph{completeness}.}
  \label{fig:faithfulness}
\end{figure}
We evaluate the faithfulness of instance concept circuits. Specifically, we measure to what extent the concept circuit can preserve the model's behavior by performing intervention experiments.

\paragraph{Metrics.}
We use two standard metrics for measuring circuit faithfulness~\cite{neel2022attribution,conmy2023towards,kramar2024atp}:
(1) \emph{Faithfulness}: measures how well a concept circuit $\mathcal{C}$ captures the behavior of the original model on a given metric $m$ ({\it e.g.,} class logits), defined as $ \frac{m(\mathcal{C}) - m(\varnothing)}{m(\mathcal{G}) - m(\varnothing)}$,
where $m(\mathcal{C})$ denotes the value of $m$ when running the model with all nodes outside of $\mathcal{C}$ mean-ablated, $\mathcal{G}$ is the full graph and $\varnothing$ is the empty graph.
(2) \emph{Completeness}: measures how necessary the circuit is by evaluating the faithfulness of $\mathcal{G}\setminus\mathcal{C}$. The lower the better.

\paragraph{Settings.}
We evaluate the faithfulness and completeness of instance concept circuits on CLIP~\cite{radford2021learning}, DINO~\cite{caron2021emerging} and supervised ViT~\cite{dosovitskiy2021an}. We randomly choose 50 classes from ImageNet~\cite{deng2009imagenet} validation set. For each class, we evaluate the instance concept circuit for this class by aggregating over all input images, and report the average metrics across all classes. We also evaluate SAE baselines with matched dictionary size and sparsity for comparison.

\paragraph{Results.}
We plot the \textit{faithfulness} and \textit{completeness} of instance concept circuits with different number of nodes (Fig.~\ref{fig:faithfulness}). We find that CLT consistently outperforms SAE on both metrics across all models and graph sparsity. In particular, on CLIP, CLT surpasses SAE by $24.4\%$ on faithfulness and $22.1\%$ on completeness in average. This suggests that CLTs capture features that are more important to original model's computation, demonstrating the effectiveness of CLTs on finding instance concept circuits.
\section{Experiments}
\begin{figure*}[t]
  \centering
  \fboxsep=0pt 
  \includegraphics[width=1.0\linewidth]{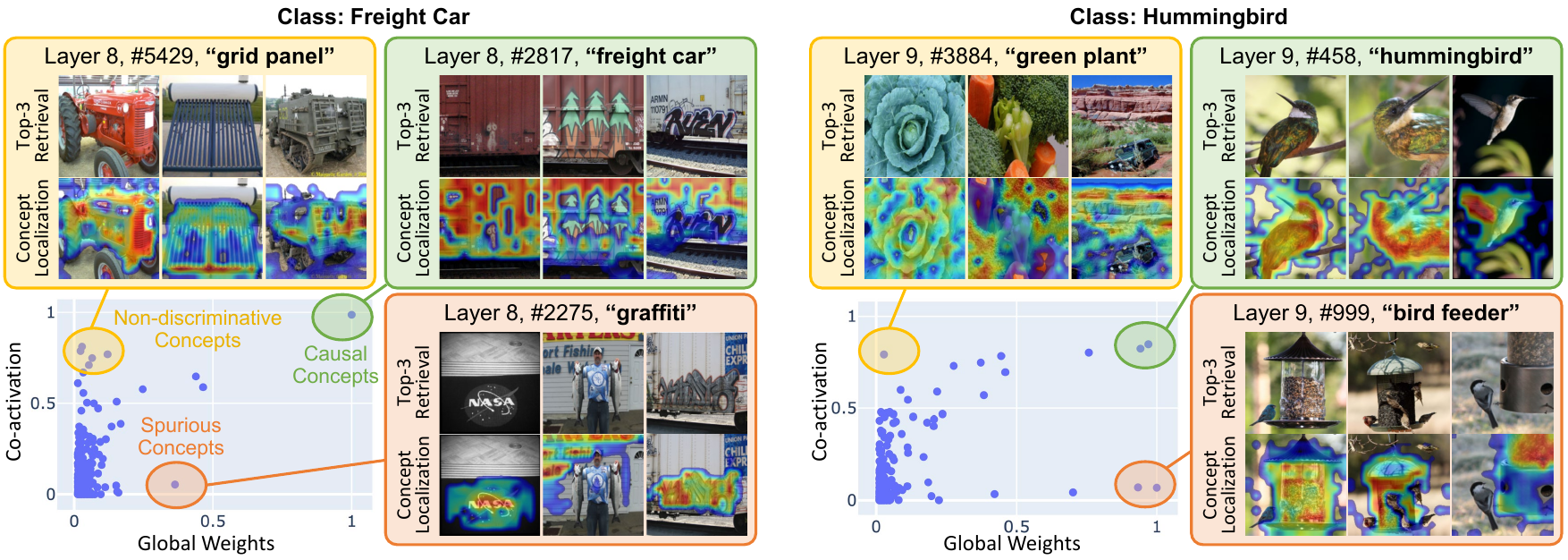}
  \caption{Automatic spurious correlation discovery on ViT-B/16. By analyzing global weights and co-activation statistics, we identify three types of concepts. \textbf{Spurious concepts (red):} concepts with high global weights but low co-activations. These concepts are spurious correlations because they rarely co-occur with the target class but are relied by the model. \textbf{Non-discriminative concepts (yellow):} concepts with high co-activations but low global weights. Although these concepts frequently co-occur with the target class, they are potentially shared by different classes, so the model doesn't rely on them to make predictions. \textbf{Causal concepts (green):} concepts with both high global weights and co-activations. These discriminative concepts exclusively co-occur with the target class so the model naturally learns to assign them high importance.}
  \label{fig:spurious_correlation_discovery}
\end{figure*}
\subsection{Automatic spurious correlation discovery}

\begin{tcolorbox}[width=\linewidth,colback={verylightgray}, colframe={lightgray},left=0pt, right=0pt, top=0pt, bottom=0pt]
{\bf Takeaway:} {our global concept circuits enable automatic spurious correlation discovery without human annotation.}
\end{tcolorbox}

\noindent Vision models have been shown to rely on spurious correlations~\cite{ye2024spurious,xiao2021noise,moayeri2022comprehensive,geirhos2018imagenet}, hindering model's generalization beyond training environments. 
Most prior work~\cite{singla2022salient,neuhaus2023spurious,plumb2022finding,abid2022meaningfully} use explainability techniques and rely on human inspections to identify spurious correlations, which is not scalable.
Different from existing methods, our global concept circuit provides relations between features encoded inside the model, enabling \emph{automatic} spurious correlation discovery.

\vspace{1em}
\noindent\textbf{Method.}
Spurious correlations are features that the model relies on but rarely co-occur with the target class in the unbiased environment. The model's reliance on a feature can be characterized by the edge weights in the global concept circuit, and the co-occurrence frequency can be captured by the feature co-activation statistics. We can thereby automatically discover spurious correlations with these two quantities.
Concretely, given a ViT model $\mathcal{F}$ and its global concept circuit $G = (\mathcal{V}, \mathcal{E})$, where nodes $v \in \mathcal{V}$ are concepts encodes in $\mathcal{F}$ and edges $(u \!\to\! v) \in \mathcal{E}$ quantify the relations between concepts, we automatically discover spurious correlations learned by $\mathcal{F}$ in two steps:
\begin{enumerate}
    \item \emph{Identify target concept.} For a target class $C$, we first identify a target concept $t$ with top-influence to the logit of $C$.
    \item \emph{Discover spurious concepts.} For target concept $t$, we calculate its image-level co-activation statistics with all source concepts on a dataset $\mathcal{D}$ of class $C$. To avoid bias from a single dataset, we combine target class images from ImageNet~\cite{deng2009imagenet}, LVIS~\cite{gupta2019lvis} and Visual Genome~\cite{krishna2017visual} dataset to construct $\mathcal{D}$. The concepts spuriously correlated with $\mathcal{C}$ are those with low co-activations but high global weights.
\end{enumerate}

\vspace{1em}
\noindent\textbf{Results.}
Fig.~\ref{fig:spurious_correlation_discovery} demonstrates example spurious correlations discovered by our method. 
Specifically, we discover spurious correlations on ImageNet pretrained ViT-B/16 for two classes: ``Freight Car'' and ``Hummingbird''. 
We find that spurious correlated concepts discovered by our method (``graffiti'' for ``Freight Car'' and ``bird feeder'' for ``Hummingbird'') align with those found by previous work~\cite{neuhaus2023spurious}. Moreover, we also identify causal concepts (frequently co-activate and model relies on) and non-discriminative concepts (frequently co-activate but model doesn't rely on), validating the effectiveness of our method.

\subsection{Spurious correlation removal}
\begin{tcolorbox}[width=\linewidth,colback={verylightgray}, colframe={lightgray},left=0pt, right=0pt, top=0pt, bottom=0pt]
{\bf Takeaway:} {CLT-based instance concept circuits exhibit superior steering capability in spurious correlation removal.}
\end{tcolorbox}

\begin{wraptable}{r}{0.45\textwidth}
    \vspace{-30pt}
    \centering
    \caption{Steering accuracy on Waterbird~\cite{sagawadistributionally} dataset. Our method achieves the best steering accuracy on the worst group (waterbirds on land, where the land background is the spurious factor).}
    \resizebox{0.9\linewidth}{!}{
        \begin{tabular}{lcc}
        \toprule
        & \begin{tabular}[c]{@{}c@{}} \textbf{Landbirds}\\ \textbf{on Land}\end{tabular} 
        & \begin{tabular}[c]{@{}c@{}} \textbf{Waterbirds}\\ \textbf{on Land}\end{tabular} \\
        \midrule
        Original & 0.98 & 0.48 \\
        SpLiCE~\cite{bhalla2024interpreting} & 0.97 & 0.60 \\
        SAE & \textbf{0.99} & 0.53 \\
        BatchTopKSAE & 0.97 & 0.56 \\
        Random & 1.00 & 0.22 \\
        Ours & 0.95 & \textbf{0.71} \\
        \bottomrule
        \end{tabular}
    }
    \label{tab:waterbird}
    \vspace{-20pt}
\end{wraptable}

\noindent Apart from discovering spurious correlations, we also want to remove spurious correlations learned by a pretrained model to make it more robust. Existing methods, such as SpLiCE~\cite{bhalla2024interpreting}, only intervene on the representations of the last layer, hindering the steering power. Although SAEs can intervene on intermediate layers, it may not faithfully capture the relations between concepts. In this section, we demonstrate CLT-based instance concept circuit's superior steering capability in spurious correlation removal.

\vspace{1em}
\noindent\textbf{Settings.}
We test the instance concept circuit's steering capability on the WaterBirds dataset~\cite{sagawadistributionally}, which spuriously correlates bird categories with background ({\it e.g.,} landbirds with land background) and results in trained classifiers performing poorly on less representative groups ({\it e.g.,} waterbirds on land background). We use CLIP-ViT-B/32 as the backbone model.

\vspace{1em}
\noindent\textbf{Method.}
Given a trained linear probe on the last layer representations of CLIP-ViT-B/32, we remove the spurious correlations with CLTs using a simple, automated strategy. Specifically, we first discover the instance concept circuit on training images with water and land background, namely $\mathcal{C}_{\text{water}}$ and $\mathcal{C}_{\text{land}}$, respectively. Intuitively, $\mathcal{C}_{\text{water}}$ and $\mathcal{C}_{\text{land}}$ both contain features for identifying bird species, which is $\mathcal{C}_{\text{water}} \cap \mathcal{C}_{\text{land}}$. We then intervene the trained linear probe by mean-ablating all features in $(\mathcal{C}_{\text{water}} \cup \mathcal{C}_{\text{land}})\setminus (\mathcal{C}_{\text{water}} \cap \mathcal{C}_{\text{land}})$, which contains background cues, when running the model.

\vspace{1em}
\noindent\textbf{Results.}
Following SpLiCE~\cite{bhalla2024interpreting}, we report accuracies on two groups: landbirds on land and waterbirds on land (worst group). 
We compare our method with SpLiCE~\cite{bhalla2024interpreting}, SAE~\cite{bricken2023monosemanticity} and BatchTopKSAE~\cite{bussmann2024batchtopk}. For SAE and BatchTopKSAE, we control the same dictionary size, sparsity, and use the same method as CLTs for fair comparison. To demonstrate that the performance gains come from correctly identifying spurious concepts, we ablate a same number of randomly selected concepts from the highly activated concepts.
As shown in Tab.~\ref{tab:waterbird}, our method significantly boost the worst group accuracy by $23\%$, while SAE only results in $5\%$ improvement. Compared to SpLiCE, which manually removes concepts related to land background ({\it e.g.,} ``bamboo'', ``forest''), our method achieves better performance with automated concept selection strategy, highlighting the concept circuit uncovers concept flows that are faithful to model behavior.

\subsection{Model comparison}
\begin{tcolorbox}[width=\linewidth,colback={verylightgray}, colframe={lightgray},left=0pt, right=0pt, top=0pt, bottom=0pt]
{\bf Takeaway:} {our global concept circuits reveal how variant supervision paradigms shape ViT's internal representation structure.}
\end{tcolorbox}

\noindent Different supervision paradigms yield noticeably different model behaviors. For example, contrastive language–image pretraining (e.g., CLIP~\cite{radford2021learning}) exhibits strong generalization ability while self-supervised objectives (e.g., DINO~\cite{caron2021emerging}) transfer well to dense prediction tasks. 
This raises a natural question: \textit{can we understand how a supervision paradigm shapes a model’s behavior by inspecting its internal mechanisms, rather than only comparing downstream accuracy}? 
Our global concept circuit provides a direct lens into this question by revealing how each model organizes and routes concept-level information flow, allowing us to relate behavioral differences (\eg, robustness or task specialization) to concrete structural differences in their global concept circuit topology.

\begin{figure*}[t]
  \centering
  \fboxsep=0pt 
  \includegraphics[width=1.0\linewidth]{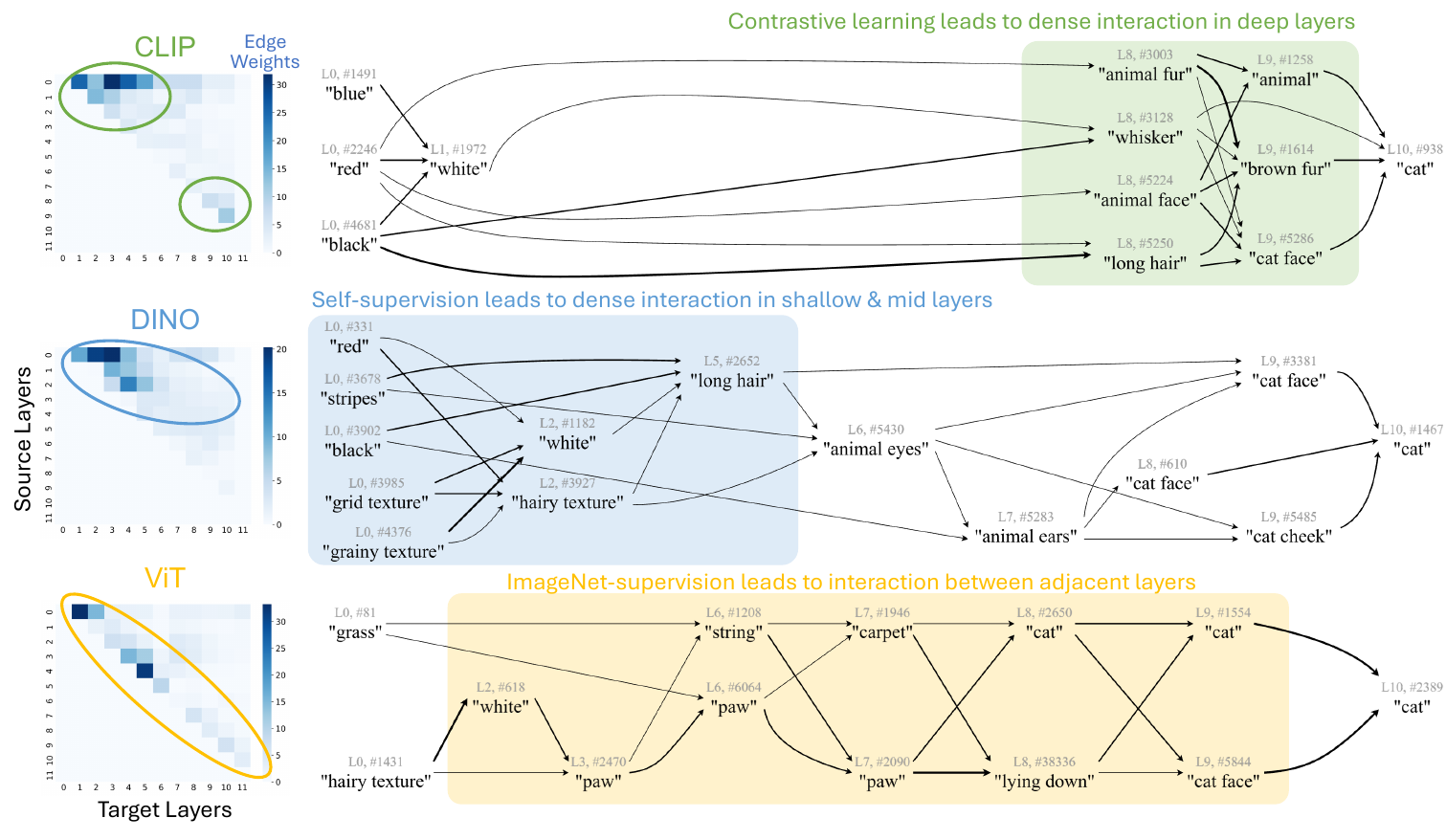}
  \caption{Comparison of global concept circuits between CLIP, DINO, and ImageNet-supervised ViT-B/16. \textbf{Left:} distributions of global edge weights aggregated by layers. \textbf{Right:} example of global concept circuits for the ``Cat'' class. Contrastively-learned CLIP exhibits dense deep layer connections, resulting in strong generalization ability; self-supervised DINO shows strong connections primarily in shallow and middle layers, thus transfers well to dense prediction tasks; ImageNet-supervised ViT has more adjacent layer interactions, leading to early development of abstract concepts.}
  \label{fig:model_comparison}
  \vspace{-10pt}
\end{figure*}
\vspace{1em}
\noindent\textbf{Method.}
The global concept circuit comprises of thousands of concepts and dense interactions between them, making it difficult to analyze. We thus aggregate the global edge weights by layers, resulting in a simplified graph whose node set is the set of layers.
This allows us to study how information flow between layers and analyze the structural differences of vision foundation models. 
In addition to the layer-level concept circuit, we also extract the subgraph of a specific class using the method described in Sec.~\ref{subsec:global_circuit} to verify our observations.

\vspace{1em}
\noindent\textbf{Results.}
Fig.~\ref{fig:model_comparison} shows the comparison between CLIP~\cite{radford2021learning}, DINO~\cite{caron2021emerging}, and ImageNet-supervised ViT-B/16~\cite{dosovitskiy2021an}. From the layer-level concept circuits and the subgraphs of the ``Cat'' class, we have the following observations: (1) CLIP has dense interactions in deep layers, with relatively weak mid-layer connections. This implies that CLIP mainly relies on high-level concepts in deep layers, aligning with prior observations that CLIP embeddings are strongly aligned with language and have strong generalization capability~\cite{radford2021learning}  (2) DINO shows strong connections between shallow and middle layers, focusing on low-level concepts such as colors and textures. This explains DINO's strong performance as a dense prediction backbone~\cite{caron2021emerging}. (3) ViT relies mainly on adjacent-layer information and shows gradual feature evolution, meaning that for easy classes, ViT can encode the information of those classes before reaching the final layer ({\it e.g.,} the ``cat'' concept in layer 8 in Fig.~\ref{fig:model_comparison}). This is consistent with prior findings that ViTs can classify many classes without using all layers~\cite{wang2021not}.
\section{Related Work}

\noindent\textbf{Interpretable machine learning (IML)}\quad
A large body of IML work explains vision models by attributing predictions to inputs or intermediate signals.
Prominent families include gradient-based saliency and class-activation methods~\cite{simonyan2013deep,zhou2016learning,selvaraju2017grad}, perturbation- and occlusion-based importance~\cite{fong2017interpretable,petsiuk2018rise,fong2019understanding}, and example-based explanations~\cite{koh2017understanding,chen2019looks,yeh2018representer}.
Other widely used tools provide local surrogate or feature-importance explanations~\cite{ribeiro2016should,lundberg2017unified}.
These approaches are often effective for highlighting \emph{where} evidence lies in an image or \emph{which} inputs matter, but they typically provide limited insight into \emph{how} internal representations compose and interact across layers, leaving much of the model’s mechanism opaque.
In contrast, our work tries to open the black box by reading interpretable concept circuits that faithfully capture the model behavior.

\vspace{1em}
\noindent\textbf{Concept-based interpretability}\quad
Concept-based methods aim to explain predictions using human-meaningful concepts rather than raw pixels or individual neurons.
Common approaches include concept bottleneck models~\cite{koh2020concept,yang2023language,oikarinen2023labelfree}, aligning latent dimensions with predefined concepts~\cite{chen2020concept,yuksekgonul2023posthoc,oikarinen2023clipdissect}, and testing sensitivity to user-defined concept directions~\cite{kim2018interpretability,yeh2020completeness}.
More recently, \emph{Sparse Autoencoders (SAEs)}~\cite{huben2023sparse,bricken2023monosemanticity,rajamanoharan2024jumping,gao2025scaling,li2026inside,ma2026medsae} have been used to decompose polysemantic activations into sparse, often more interpretable, \emph{model-native} features, enabling downstream analyses such as feature-level attribution and circuit discovery~\cite{neel2022attribution,conmy2023towards,kramar2024atp}.
However, SAE-based pipelines typically emphasize \emph{instance-dependent} behavior~\cite{dunefsky2024transcoders} and recovering a reliable global picture of feature interactions often requires expensive dataset-scale attribution and aggregation.
In contrast, our approach uses CLTs to recover a \emph{concept graph} directly from learned, input-invariant weights, while still supporting instance-level circuit extraction.

\vspace{1em}
\noindent\textbf{Mechanistic interpretability}\quad
Mechanistic interpretability seeks to reverse engineer neural networks into human-understandable algorithms by analyzing weights, features, and circuits~\cite{elhage2021mathematical,olsson2022context,conmy2023towards,nanda2023progress,bereska2024mechanistic}.
Early work on circuits in vision and language models demonstrated that specific behaviors can be implemented by sparse computational subgraphs~\cite{olah2020zoom,cammarata2021curve}, and more recent methods automate circuit discovery via attribution patching and related techniques~\cite{kramar2024atp,marks2025sparse}.
Our work follows this bottom-up perspective, but focuses on extracting \emph{concept-level} circuits in ViTs using CLTs, yielding both instance  and global concept circuits that can be used for auditing shortcuts and comparing models.
\section{Conclusion}
We presented a CLT-based framework for reading \emph{concept circuits} from Vision Transformers, turning the residual stream into sparse, interpretable concepts and exposing how these concepts interact across layers. Central to our approach is a two-level view: a \emph{global} concept circuit that summarizes input-invariant concept-to-concept structure encoded in the model’s parameters, and \emph{instance} concept circuits that reveal the active concepts and pathways used for a particular prediction. We showed that this representation is not only interpretable but also useful: global circuits enable scalable auditing and principled comparisons across models trained with different supervision paradigms, while instance circuits support fine-grained diagnosis and targeted interventions.

\paragraph{Limitations:} CLTs primarily capture MLP pathways and only indirectly reflect attention structure, concept interpretation requires human inspection, and our global graphs do not fully capture causal relations.

\section*{Acknowledgements}
This work is supported by the National Science Foundation under grant numbers CAREER 2340074, SLES 2416937, III CORE 2412675 and National Institutes of Health under grant number R21CA301093. Any opinions, findings and conclusions or recommendations expressed in this material are those of the authors and do not reflect the views of the supporting entities.

%
%
\bibliographystyle{splncs04}
\bibliography{main}

\begin{thebibliography}{10}
\providecommand{\url}[1]{\texttt{#1}}
\providecommand{\urlprefix}{URL }
\providecommand{\doi}[1]{https://doi.org/#1}

\bibitem{abid2022meaningfully}
Abid, A., Yuksekgonul, M., Zou, J.: Meaningfully debugging model mistakes using conceptual counterfactual explanations. In: International Conference on Machine Learning. pp. 66--88. PMLR (2022)

\bibitem{ameisen2025circuit}
Ameisen, E., Lindsey, J., Pearce, A., Gurnee, W., Turner, N.L., Chen, B., Citro, C., Abrahams, D., Carter, S., Hosmer, B., Marcus, J., Sklar, M., Templeton, A., Bricken, T., McDougall, C., Cunningham, H., Henighan, T., Jermyn, A., Jones, A., Persic, A., Qi, Z., Ben~Thompson, T., Zimmerman, S., Rivoire, K., Conerly, T., Olah, C., Batson, J.: Circuit tracing: Revealing computational graphs in language models. Transformer Circuits Thread  (2025), \url{https://transformer-circuits.pub/2025/attribution-graphs/methods.html}

\bibitem{bereska2024mechanistic}
Bereska, L., Gavves, S.: Mechanistic interpretability for {AI} safety - a review. Transactions on Machine Learning Research  (2024), \url{https://openreview.net/forum?id=ePUVetPKu6}, survey Certification, Expert Certification

\bibitem{bhalla2024interpreting}
Bhalla, U., Oesterling, A., Srinivas, S., Calmon, F., Lakkaraju, H.: Interpreting clip with sparse linear concept embeddings (splice). Advances in Neural Information Processing Systems  \textbf{37},  84298--84328 (2024)

\bibitem{bricken2023monosemanticity}
Bricken, T., Templeton, A., Batson, J., Chen, B., Jermyn, A., Conerly, T., Turner, N., Anil, C., Denison, C., Askell, A., Lasenby, R., Wu, Y., Kravec, S., Schiefer, N., Maxwell, T., Joseph, N., Hatfield-Dodds, Z., Tamkin, A., Nguyen, K., McLean, B., Burke, J.E., Hume, T., Carter, S., Henighan, T., Olah, C.: Towards monosemanticity: Decomposing language models with dictionary learning. Transformer Circuits Thread  (2023), https://transformer-circuits.pub/2023/monosemantic-features/index.html

\bibitem{bussmann2024batchtopk}
Bussmann, B., Leask, P., Nanda, N.: Batchtopk sparse autoencoders. arXiv preprint arXiv:2412.06410  (2024)

\bibitem{cammarata2021curve}
Cammarata, N., Goh, G., Carter, S., Voss, C., Schubert, L., Olah, C.: Curve circuits. Distill  (2021). \doi{10.23915/distill.00024.006}, https://distill.pub/2020/circuits/curve-circuits

\bibitem{caron2021emerging}
Caron, M., Touvron, H., Misra, I., J{\'e}gou, H., Mairal, J., Bojanowski, P., Joulin, A.: Emerging properties in self-supervised vision transformers. In: Proceedings of the IEEE/CVF international conference on computer vision. pp. 9650--9660 (2021)

\bibitem{chen2019looks}
Chen, C., Li, O., Tao, D., Barnett, A., Rudin, C., Su, J.K.: This looks like that: deep learning for interpretable image recognition. Advances in neural information processing systems  \textbf{32} (2019)

\bibitem{chen2023vision}
Chen, Z., Duan, Y., Wang, W., He, J., Lu, T., Dai, J., Qiao, Y.: Vision transformer adapter for dense predictions. In: The Eleventh International Conference on Learning Representations (2023), \url{https://openreview.net/forum?id=plKu2GByCNW}

\bibitem{chen2020concept}
Chen, Z., Bei, Y., Rudin, C.: Concept whitening for interpretable image recognition. Nature Machine Intelligence  \textbf{2}(12),  772--782 (2020)

\bibitem{conmy2023towards}
Conmy, A., Mavor-Parker, A., Lynch, A., Heimersheim, S., Garriga-Alonso, A.: Towards automated circuit discovery for mechanistic interpretability. Advances in Neural Information Processing Systems  \textbf{36},  16318--16352 (2023)

\bibitem{deng2009imagenet}
Deng, J., Dong, W., Socher, R., Li, L.J., Li, K., Fei-Fei, L.: Imagenet: A large-scale hierarchical image database. In: 2009 IEEE conference on computer vision and pattern recognition. pp. 248--255. Ieee (2009)

\bibitem{dosovitskiy2021an}
Dosovitskiy, A., Beyer, L., Kolesnikov, A., Weissenborn, D., Zhai, X., Unterthiner, T., Dehghani, M., Minderer, M., Heigold, G., Gelly, S., Uszkoreit, J., Houlsby, N.: An image is worth 16x16 words: Transformers for image recognition at scale. In: International Conference on Learning Representations (2021), \url{https://openreview.net/forum?id=YicbFdNTTy}

\bibitem{dunefsky2024transcoders}
Dunefsky, J., Chlenski, P., Nanda, N.: Transcoders find interpretable llm feature circuits. Advances in Neural Information Processing Systems  \textbf{37},  24375--24410 (2024)

\bibitem{elhage2021mathematical}
Elhage, N., Nanda, N., Olsson, C., Henighan, T., Joseph, N., Mann, B., Askell, A., Bai, Y., Chen, A., Conerly, T., DasSarma, N., Drain, D., Ganguli, D., Hatfield-Dodds, Z., Hernandez, D., Jones, A., Kernion, J., Lovitt, L., Ndousse, K., Amodei, D., Brown, T., Clark, J., Kaplan, J., McCandlish, S., Olah, C.: A mathematical framework for transformer circuits. Transformer Circuits Thread  (2021), https://transformer-circuits.pub/2021/framework/index.html

\bibitem{fong2019understanding}
Fong, R., Patrick, M., Vedaldi, A.: Understanding deep networks via extremal perturbations and smooth masks. In: Proceedings of the IEEE/CVF international conference on computer vision. pp. 2950--2958 (2019)

\bibitem{fong2017interpretable}
Fong, R.C., Vedaldi, A.: Interpretable explanations of black boxes by meaningful perturbation. In: Proceedings of the IEEE international conference on computer vision. pp. 3429--3437 (2017)

\bibitem{gao2025scaling}
Gao, L., la~Tour, T.D., Tillman, H., Goh, G., Troll, R., Radford, A., Sutskever, I., Leike, J., Wu, J.: Scaling and evaluating sparse autoencoders. In: The Thirteenth International Conference on Learning Representations (2025), \url{https://openreview.net/forum?id=tcsZt9ZNKD}

\bibitem{geirhos2018imagenet}
Geirhos, R., Rubisch, P., Michaelis, C., Bethge, M., Wichmann, F.A., Brendel, W.: Imagenet-trained cnns are biased towards texture; increasing shape bias improves accuracy and robustness. In: International conference on learning representations (2018)

\bibitem{gupta2019lvis}
Gupta, A., Dollar, P., Girshick, R.: Lvis: A dataset for large vocabulary instance segmentation. In: Proceedings of the IEEE/CVF conference on computer vision and pattern recognition. pp. 5356--5364 (2019)

\bibitem{huben2023sparse}
Huben, R., Cunningham, H., Smith, L.R., Ewart, A., Sharkey, L.: Sparse autoencoders find highly interpretable features in language models. In: The Twelfth International Conference on Learning Representations (2023)

\bibitem{kim2018interpretability}
Kim, B., Wattenberg, M., Gilmer, J., Cai, C., Wexler, J., Viegas, F., et~al.: Interpretability beyond feature attribution: Quantitative testing with concept activation vectors (tcav). In: International conference on machine learning. pp. 2668--2677. PMLR (2018)

\bibitem{koh2017understanding}
Koh, P.W., Liang, P.: Understanding black-box predictions via influence functions. In: International conference on machine learning. pp. 1885--1894. PMLR (2017)

\bibitem{koh2020concept}
Koh, P.W., Nguyen, T., Tang, Y.S., Mussmann, S., Pierson, E., Kim, B., Liang, P.: Concept bottleneck models. In: International conference on machine learning. pp. 5338--5348. PMLR (2020)

\bibitem{kramar2024atp}
Kram{\'a}r, J., Lieberum, T., Shah, R., Nanda, N.: Atp*: An efficient and scalable method for localizing llm behaviour to components. arXiv preprint arXiv:2403.00745  (2024)

\bibitem{krishna2017visual}
Krishna, R., Zhu, Y., Groth, O., Johnson, J., Hata, K., Kravitz, J., Chen, S., Kalantidis, Y., Li, L.J., Shamma, D.A., et~al.: Visual genome: Connecting language and vision using crowdsourced dense image annotations. International journal of computer vision  \textbf{123}(1),  32--73 (2017)

\bibitem{li2026inside}
Li, T., Chen, Y., Ma, M., Peng, X.: Inside the visual mind: Neuroscience-motivated concept circuits for interpreting and steering vision transformers. In: Forty-third International Conference on Machine Learning (2026), \url{https://openreview.net/forum?id=P2dl32LtuQ}

\bibitem{lundberg2017unified}
Lundberg, S.M., Lee, S.I.: A unified approach to interpreting model predictions. Advances in neural information processing systems  \textbf{30} (2017)

\bibitem{ma2026medsae}
Ma, M., Peng, Y., Li, T., Lin, L., Beylergil, V., Zhao, B., Akin, O., Peng, X.: {Medical AI Encodes a 'Feeling of Error': Verifying Cancer Segmentation via Internal Concepts}. In: Proceedings of the European Conference on Computer Vision (ECCV) (2026)

\bibitem{marks2025sparse}
Marks, S., Rager, C., Michaud, E.J., Belinkov, Y., Bau, D., Mueller, A.: Sparse feature circuits: Discovering and editing interpretable causal graphs in language models. In: The Thirteenth International Conference on Learning Representations (2025), \url{https://openreview.net/forum?id=I4e82CIDxv}

\bibitem{moayeri2022comprehensive}
Moayeri, M., Pope, P., Balaji, Y., Feizi, S.: A comprehensive study of image classification model sensitivity to foregrounds, backgrounds, and visual attributes. In: Proceedings of the IEEE/CVF Conference on Computer Vision and Pattern Recognition. pp. 19087--19097 (2022)

\bibitem{neel2022attribution}
Nanda, N.: Attribution patching: Activation patching at industrial scale  (2022), \url{https://www.neelnanda.io/mechanistic-interpretability/attribution-patching}

\bibitem{nanda2023progress}
Nanda, N., Chan, L., Lieberum, T., Smith, J., Steinhardt, J.: Progress measures for grokking via mechanistic interpretability. In: The Eleventh International Conference on Learning Representations (2023), \url{https://openreview.net/forum?id=9XFSbDPmdW}

\bibitem{neuhaus2023spurious}
Neuhaus, Y., Augustin, M., Boreiko, V., Hein, M.: Spurious features everywhere-large-scale detection of harmful spurious features in imagenet. In: Proceedings of the IEEE/CVF International Conference on Computer Vision. pp. 20235--20246 (2023)

\bibitem{oikarinen2023labelfree}
Oikarinen, T., Das, S., Nguyen, L.M., Weng, T.W.: Label-free concept bottleneck models. In: The Eleventh International Conference on Learning Representations (2023), \url{https://openreview.net/forum?id=FlCg47MNvBA}

\bibitem{oikarinen2023clipdissect}
Oikarinen, T., Weng, T.W.: {CLIP}-dissect: Automatic description of neuron representations in deep vision networks. In: The Eleventh International Conference on Learning Representations (2023), \url{https://openreview.net/forum?id=iPWiwWHc1V}

\bibitem{olah2020zoom}
Olah, C., Cammarata, N., Schubert, L., Goh, G., Petrov, M., Carter, S.: Zoom in: An introduction to circuits. Distill  (2020). \doi{10.23915/distill.00024.001}, https://distill.pub/2020/circuits/zoom-in

\bibitem{olsson2022context}
Olsson, C., Elhage, N., Nanda, N., Joseph, N., DasSarma, N., Henighan, T., Mann, B., Askell, A., Bai, Y., Chen, A., Conerly, T., Drain, D., Ganguli, D., Hatfield-Dodds, Z., Hernandez, D., Johnston, S., Jones, A., Kernion, J., Lovitt, L., Ndousse, K., Amodei, D., Brown, T., Clark, J., Kaplan, J., McCandlish, S., Olah, C.: In-context learning and induction heads. Transformer Circuits Thread  (2022), https://transformer-circuits.pub/2022/in-context-learning-and-induction-heads/index.html

\bibitem{pach2025sparse}
Pach, M., Karthik, S., Bouniot, Q., Belongie, S., Akata, Z.: Sparse autoencoders learn monosemantic features in vision-language models. In: The Thirty-ninth Annual Conference on Neural Information Processing Systems (2025), \url{https://openreview.net/forum?id=DaNnkQJSQf}

\bibitem{peng2026inside}
Peng, Y., Ma, M., Yao, Z., Peng, X.: Inside-out: Measuring generalization in vision transformers through inner workings. In: Proceedings of the IEEE/CVF Conference on Computer Vision and Pattern Recognition. pp. 38936--38946 (2026)

\bibitem{petsiuk2018rise}
Petsiuk, V., Das, A., Saenko, K.: Rise: Randomized input sampling for explanation of black-box models. arXiv preprint arXiv:1806.07421  (2018)

\bibitem{plumb2022finding}
Plumb, G., Ribeiro, M.T., Talwalkar, A.: Finding and fixing spurious patterns with explanations. Transactions on Machine Learning Research  (2022), \url{https://openreview.net/forum?id=whJPugmP5I}, expert Certification

\bibitem{radford2021learning}
Radford, A., Kim, J.W., Hallacy, C., Ramesh, A., Goh, G., Agarwal, S., Sastry, G., Askell, A., Mishkin, P., Clark, J., et~al.: Learning transferable visual models from natural language supervision. In: International conference on machine learning. pp. 8748--8763. PmLR (2021)

\bibitem{rajamanoharan2024jumping}
Rajamanoharan, S., Lieberum, T., Sonnerat, N., Conmy, A., Varma, V., Kram{\'a}r, J., Nanda, N.: Jumping ahead: Improving reconstruction fidelity with jumprelu sparse autoencoders. arXiv preprint arXiv:2407.14435  (2024)

\bibitem{ribeiro2016should}
Ribeiro, M.T., Singh, S., Guestrin, C.: " why should i trust you?" explaining the predictions of any classifier. In: Proceedings of the 22nd ACM SIGKDD international conference on knowledge discovery and data mining. pp. 1135--1144 (2016)

\bibitem{sagawadistributionally}
Sagawa, S., Koh, P.W., Hashimoto, T.B., Liang, P.: Distributionally robust neural networks. In: International Conference on Learning Representations (2019)

\bibitem{selvaraju2017grad}
Selvaraju, R.R., Cogswell, M., Das, A., Vedantam, R., Parikh, D., Batra, D.: Grad-cam: Visual explanations from deep networks via gradient-based localization. In: Proceedings of the IEEE international conference on computer vision. pp. 618--626 (2017)

\bibitem{simonyan2013deep}
Simonyan, K., Vedaldi, A., Zisserman, A.: Deep inside convolutional networks: Visualising image classification models and saliency maps. arXiv preprint arXiv:1312.6034  (2013)

\bibitem{singla2022salient}
Singla, S., Feizi, S.: Salient imagenet: How to discover spurious features in deep learning? In: International Conference on Learning Representations (2022), \url{https://openreview.net/forum?id=XVPqLyNxSyh}

\bibitem{wang2021not}
Wang, Y., Huang, R., Song, S., Huang, Z., Huang, G.: Not all images are worth 16x16 words: Dynamic transformers for efficient image recognition. Advances in neural information processing systems  \textbf{34},  11960--11973 (2021)

\bibitem{wu2024clipself}
Wu, S., Zhang, W., Xu, L., Jin, S., Li, X., Liu, W., Loy, C.C.: {CLIPS}elf: Vision transformer distills itself for open-vocabulary dense prediction. In: The Twelfth International Conference on Learning Representations (2024), \url{https://openreview.net/forum?id=DjzvJCRsVf}

\bibitem{xiao2021noise}
Xiao, K.Y., Engstrom, L., Ilyas, A., Madry, A.: Noise or signal: The role of image backgrounds in object recognition. In: International Conference on Learning Representations (2021), \url{https://openreview.net/forum?id=gl3D-xY7wLq}

\bibitem{yang2023language}
Yang, Y., Panagopoulou, A., Zhou, S., Jin, D., Callison-Burch, C., Yatskar, M.: Language in a bottle: Language model guided concept bottlenecks for interpretable image classification. In: Proceedings of the IEEE/CVF conference on computer vision and pattern recognition. pp. 19187--19197 (2023)

\bibitem{ye2024spurious}
Ye, W., Zheng, G., Cao, X., Ma, Y., Zhang, A.: Spurious correlations in machine learning: A survey. arXiv preprint arXiv:2402.12715  (2024)

\bibitem{yeh2020completeness}
Yeh, C.K., Kim, B., Arik, S., Li, C.L., Pfister, T., Ravikumar, P.: On completeness-aware concept-based explanations in deep neural networks. Advances in neural information processing systems  \textbf{33},  20554--20565 (2020)

\bibitem{yeh2018representer}
Yeh, C.K., Kim, J., Yen, I.E.H., Ravikumar, P.K.: Representer point selection for explaining deep neural networks. Advances in neural information processing systems  \textbf{31} (2018)

\bibitem{yuksekgonul2023posthoc}
Yuksekgonul, M., Wang, M., Zou, J.: Post-hoc concept bottleneck models. In: The Eleventh International Conference on Learning Representations (2023), \url{https://openreview.net/forum?id=nA5AZ8CEyow}

\bibitem{zhou2016learning}
Zhou, B., Khosla, A., Lapedriza, A., Oliva, A., Torralba, A.: Learning deep features for discriminative localization. In: Proceedings of the IEEE conference on computer vision and pattern recognition. pp. 2921--2929 (2016)

\end{thebibliography}

\newpage

\appendix

\makeappendixtitle{%
Supplementary Material for\\[0.2em]
``World Knowledge'' in the Weights: Reading Concept Circuits of Vision Transformers
}
\vspace{-20pt}

\setcounter{figure}{7}
\setcounter{table}{2}

\section{Details of Cross-layer Transcoders}
Following Ameison et al.~\cite{ameisen2025circuit}, we use JumpReLU~\cite{rajamanoharan2024jumping} as the activation function of CLTs, which can be expressed as:
\begin{equation}
\mathrm{JumpReLU}(\mathbf{h}^l) = \begin{cases}
\mathbf{h}^l, & \mathbf{h}^l > \tau \\
0, & \text{otherwise}
\end{cases},
\end{equation}
where $\mathbf{h}^l=\mathbf{W}_{\mathrm{enc}}^l\mathbf{x}^l$ is the pre-activations of layer $l$ and $\tau$ is a learnable threshold. To optimize the weights of CLT, we minimize the following loss:
\begin{equation}
\mathcal{L}_{\mathrm{CLT}}=\sum_{l=1}^L\|\mathbf{y}^l-\hat{\mathbf{y}}^l\|_2^2+\lambda_1\mathcal{L}_{\mathrm{spa}}+\lambda_2\mathcal{L}_{\mathrm{preact}}.
\end{equation}
$\mathcal{L}_{\mathrm{spa}}$ is the sparsity penalty:
\begin{equation}
\mathcal{L}_{\mathrm{spa}}=\sum_{l=1}^L\sum_{i=1}^f\mathrm{tanh}(c\cdot\|\mathbf{W}_\mathrm{dec,i}^l\|\cdot z_i^l),
\end{equation}
where $f$ is the dictionary size, $\mathbf{W}_\mathrm{dec,i}^l$ is the concatenation of all decoder vectors of the $i$-th feature in layer $l$ and $c$ is a hyper-parameter. $\mathcal{L}_{\mathrm{preact}}$ is the pre-activation loss for preventing dead neurons:
\begin{equation}
\mathcal{L}_{\mathrm{preact}}=\sum_{l=1}^L\sum_{i=1}^f(-h_i^l),
\end{equation}
where $h_i^l$ is the pre-activation of the $i$-th feature in layer $l$.

\section{Implementation Details}
\label{supp:impl_detail}
We train the CLT using the Adam optimizer with a learning rate of $5\times 10^{-5}$, $\beta_1=0.9$, $\beta_2=0.999$. The model is trained on ImageNet training set for $250,000,000$ tokens with a batch size of $4096$. During training, the sparsity penalty $\lambda$ is linearly ramped from $0$ to its final value, and the learning rate is linearly decayed to $0$ for the final $20\%$ of the training steps. We train CLTs on the image tokens and \texttt{CLS} tokens of CLIP-ViT-B/32, CLIP-ViT-B/16, ImageNet supervised ViT-B/16 and DINO-ViT-B/16. The activations scaled to have an average norm of $\sqrt{d_{model}}$ in each layer. The scaling factors are estimated using the first $1000$ batches. We set $c$ to $4$ and $\lambda_2$ to $3\times10^{-5}$. We initialize the JumpReLU threshold $\tau$ to $0.03$. We sweep the hyper-parameters $\lambda_1$ over $[1,2,3,4,5]$ and dictionary size over $[1536,3072,6144,12288]$ on CLIP-ViT-B/32, and choose the best hyper-parameters for all models.

\section{Additional Visualizations}
\begin{figure*}[t]
  \centering
  \fboxsep=0pt 
  \includegraphics[width=0.78\linewidth]{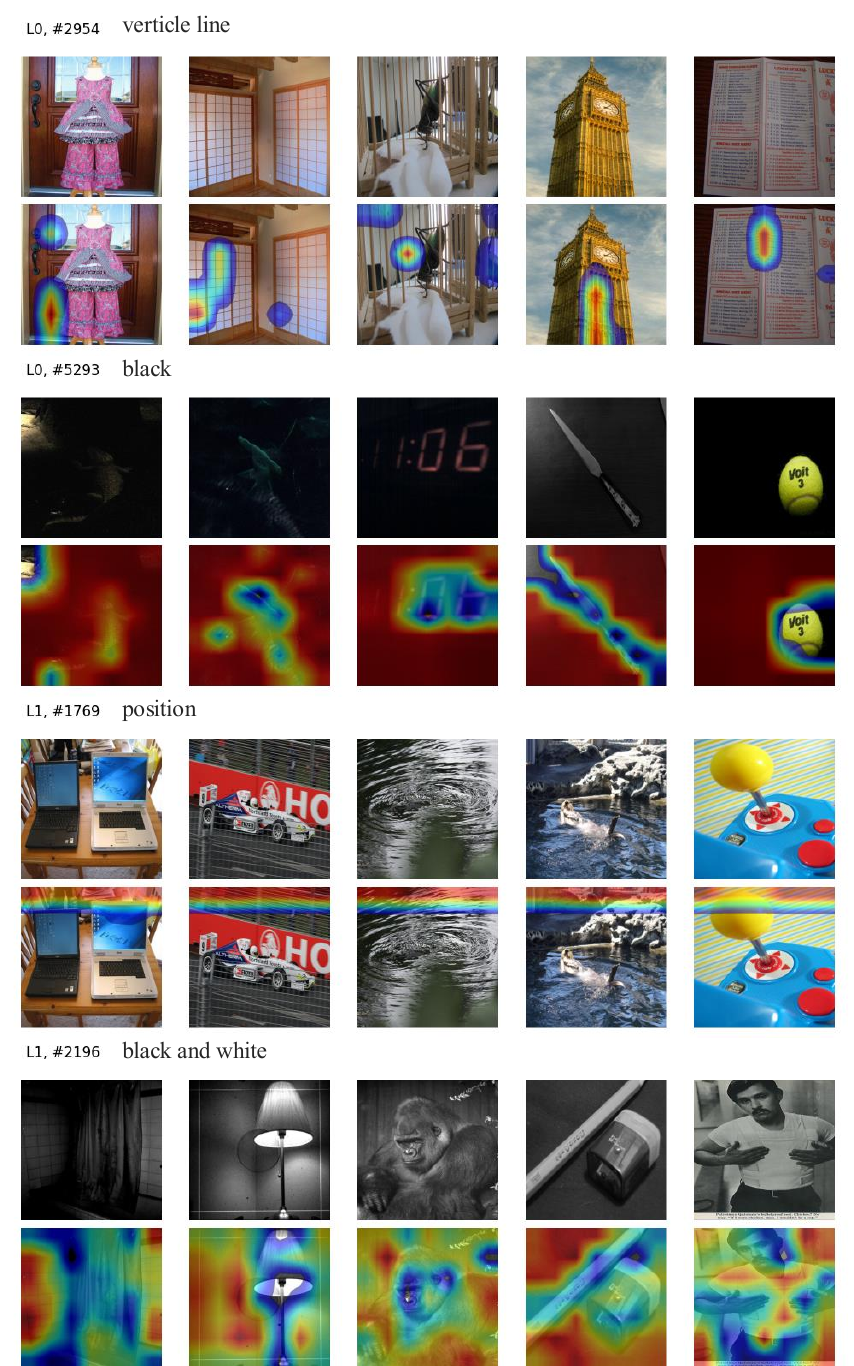}
  \caption{CLIP-ViT-B/32 layer 0 and 1 features.}
  \label{fig:supp_1}
\end{figure*}

\begin{figure*}[t]
  \centering
  \fboxsep=0pt 
  \includegraphics[width=0.78\linewidth]{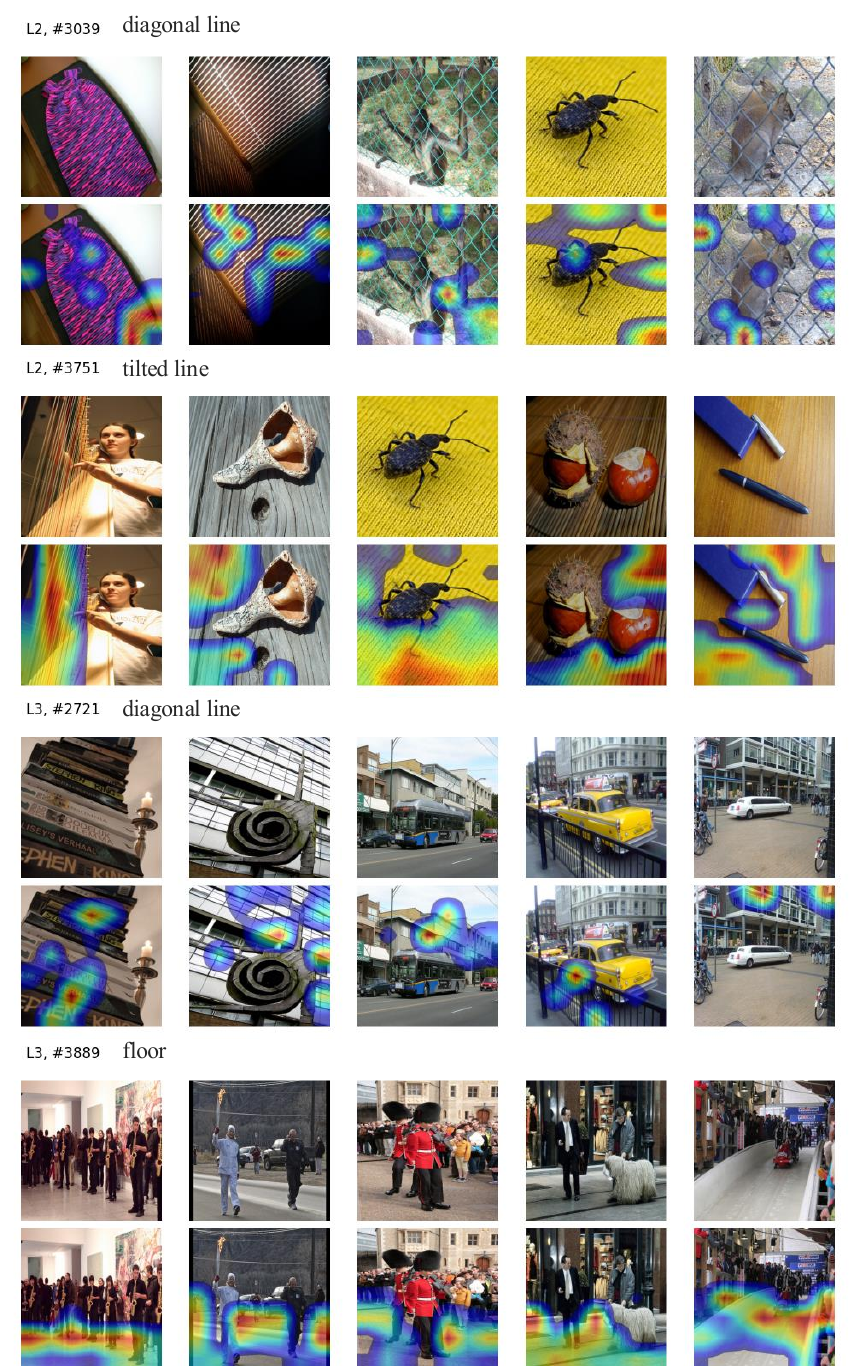}
  \caption{CLIP-ViT-B/32 layer 2 and 3 features.}
  \label{fig:supp_2}
\end{figure*}

\begin{figure*}[t]
  \centering
  \fboxsep=0pt 
  \includegraphics[width=0.78\linewidth]{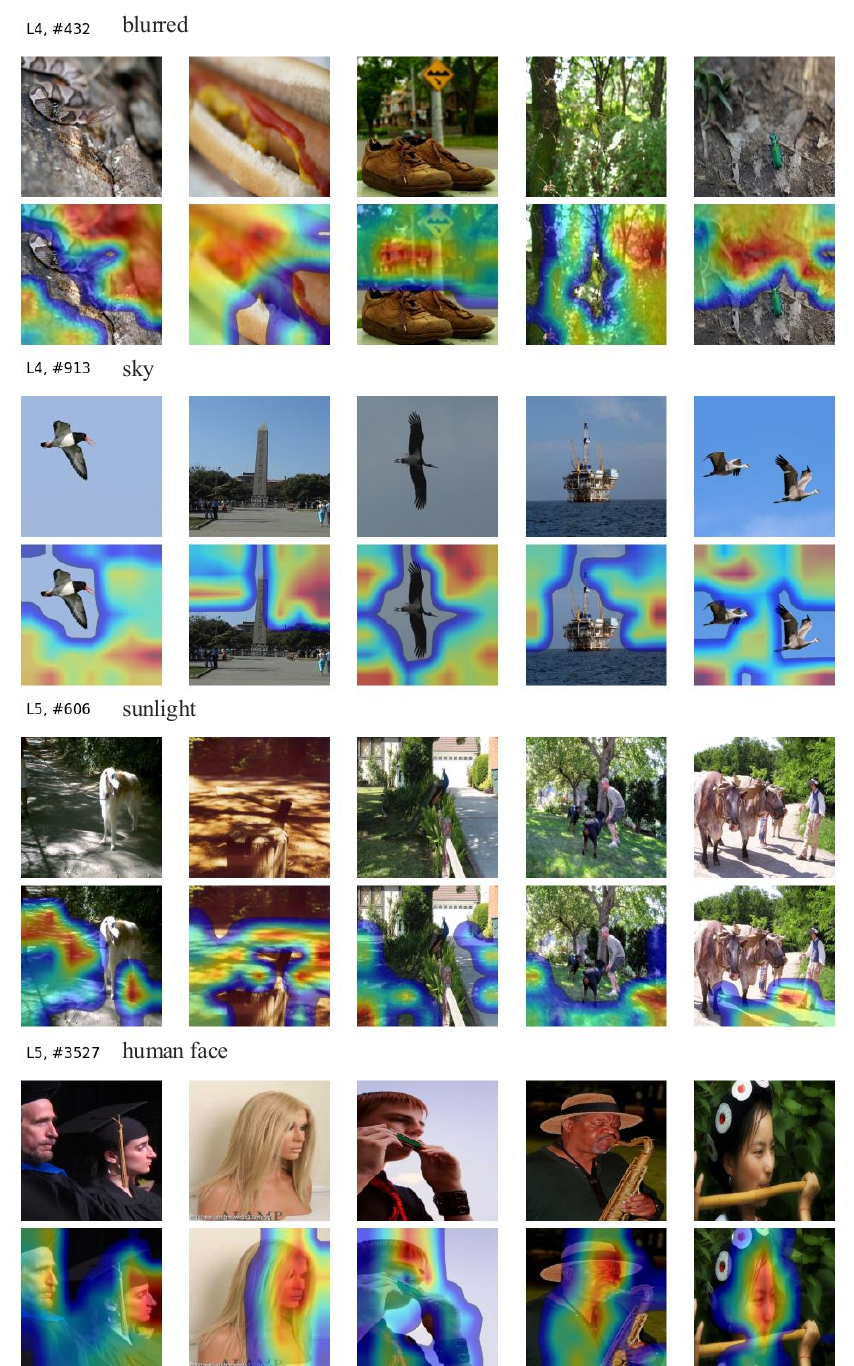}
  \caption{CLIP-ViT-B/32 layer 4 and 5 features.}
  \label{fig:supp_3}
\end{figure*}

\begin{figure*}[t]
  \centering
  \fboxsep=0pt 
  \includegraphics[width=0.78\linewidth]{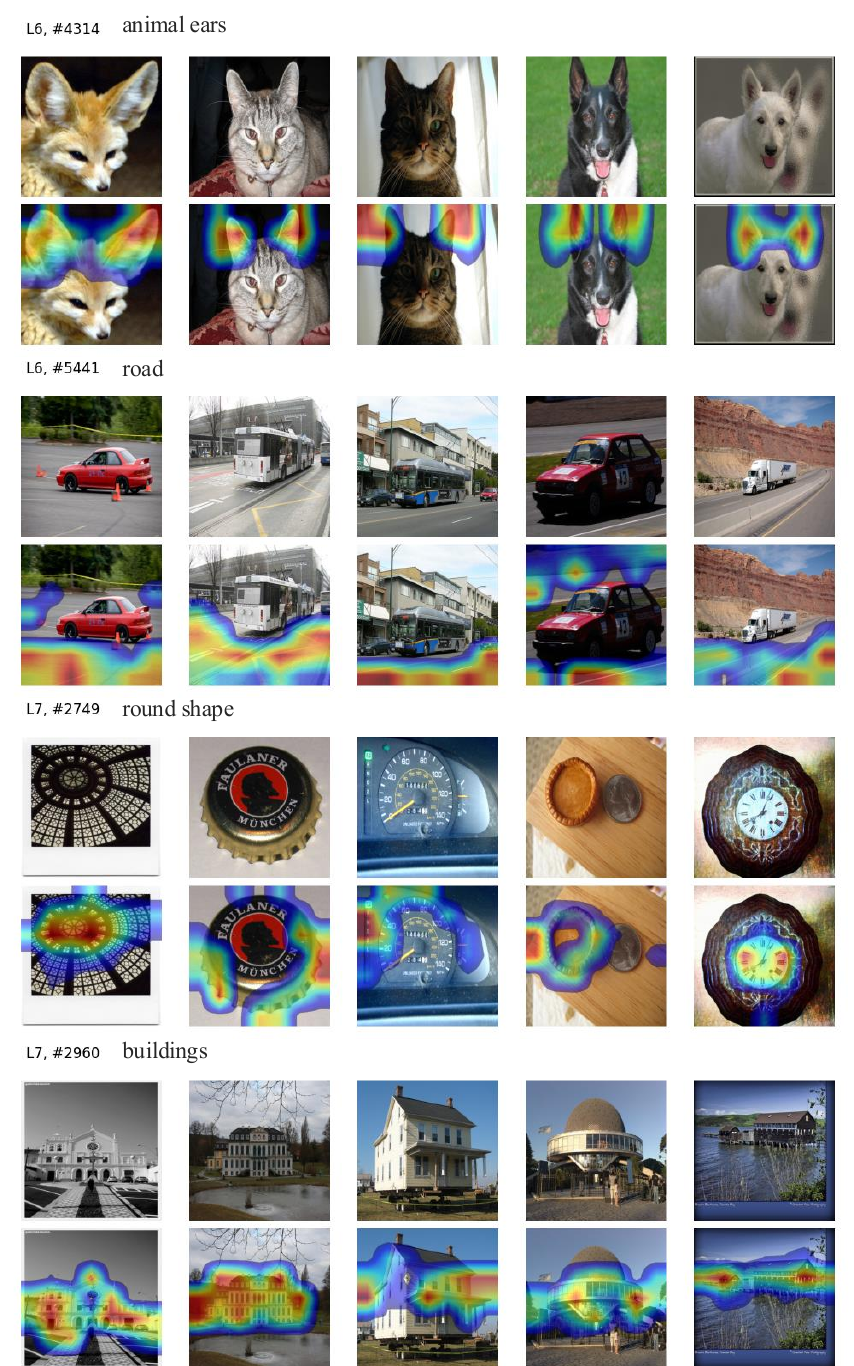}
  \caption{CLIP-ViT-B/32 layer 6 and 7 features.}
  \label{fig:supp_4}
\end{figure*}

\begin{figure*}[t]
  \centering
  \fboxsep=0pt 
  \includegraphics[width=0.78\linewidth]{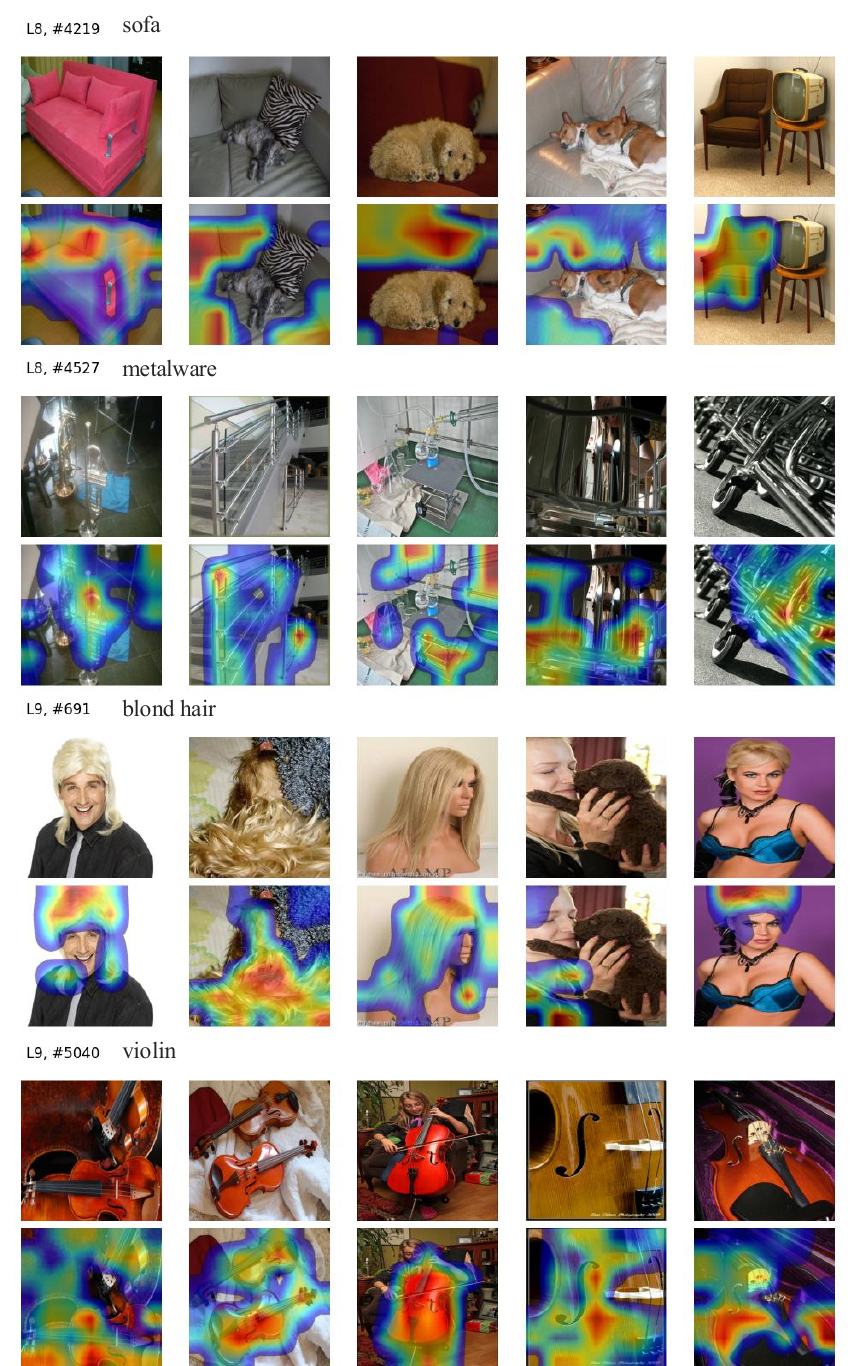}
  \caption{CLIP-ViT-B/32 layer 8 and 9 features.}
  \label{fig:supp_5}
\end{figure*}

\begin{figure*}[t]
  \centering
  \fboxsep=0pt 
  \includegraphics[width=0.78\linewidth]{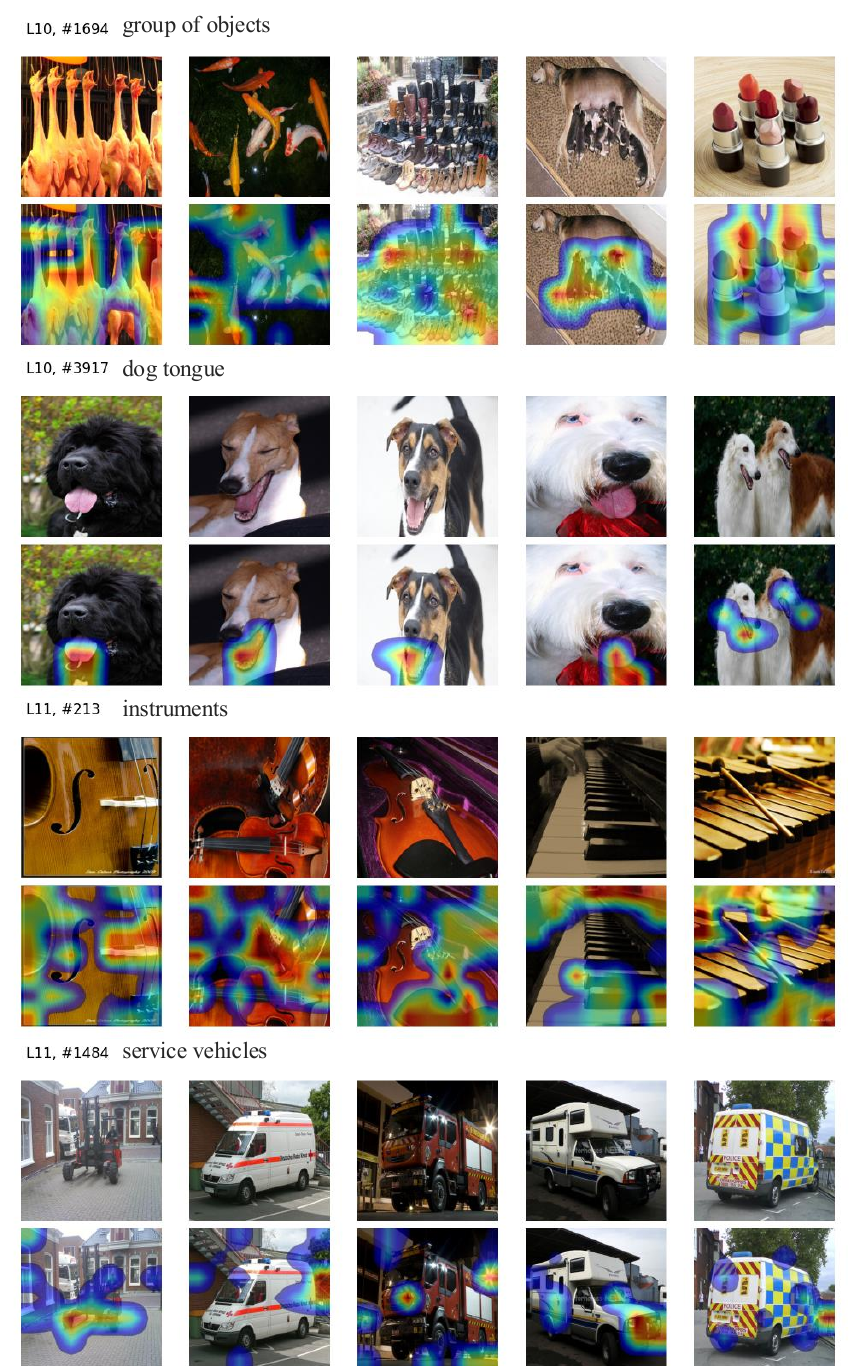}
  \caption{CLIP-ViT-B/32 layer 10 and 11 features.}
  \label{fig:supp_6}
\end{figure*}

We provide additional feature visualizations of each layer for the CLT trained on CLIP-ViT-B/32. The concepts evolve from color and line (Fig.~\ref{fig:supp_1}, Fig.~\ref{fig:supp_2}) to shape and part (Fig.~\ref{fig:supp_3}, Fig.~\ref{fig:supp_4}) and finally to object and abstract concepts (Fig.~\ref{fig:supp_5}, Fig.~\ref{fig:supp_6}).

\FloatBarrier
\end{document}